\documentclass[sigconf]{acmart}
\pdfoutput=1
\AtBeginDocument{%
  }

\definecolor{myblue}{rgb}{0.85, 0.85, 1} 
\definecolor{mygreen}{rgb}{0.8, 1, 0.8}
\definecolor{myred}{rgb}{1, 0.85, 0.85} 
\definecolor{mygray}{rgb}{0.98, 0.98, 0.98}
\definecolor{skyblue}{RGB}{135,206,235} 

\usepackage{xcolor}

\usepackage{hyperref}

\usepackage{url}
\usepackage{graphicx}
\usepackage{booktabs}
\usepackage{multirow}

\usepackage{algorithm}
\usepackage{algorithmic}
\usepackage{subcaption}
\usepackage{amsmath}
\usepackage{pifont}
\usepackage{tabularx}
\usepackage{float}
\usepackage{wrapfig}
\usepackage{colortbl}
\usepackage{enumitem}
\usepackage{fancyhdr}

\begin{document}

\copyrightyear{2025}
\acmYear{2025}
\setcopyright{acmlicensed}
\acmConference[KDD '25] {Proceedings of the 31st ACM SIGKDD Conference on Knowledge Discovery and Data Mining V.2}{August 3--7, 2025}{Toronto, ON, Canada.}
\acmBooktitle{Proceedings of the 31st ACM SIGKDD Conference on Knowledge Discovery and Data Mining V.2 (KDD '25), August 3--7, 2025, Toronto, ON, Canada}
\acmISBN{979-8-4007-1454-2/25/08}
\acmDOI{10.1145/3711896.3736901}

\settopmatter{printacmref=true}

\title{%
  \raisebox{-0.3ex}{\includegraphics[height=2ex]{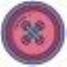}}%
  Cuff-KT: Tackling Learners' Real-time Learning Pattern\\
  Adjustment via Tuning-Free Knowledge State Guided Model Updating
}

\author{Yiyun Zhou}
\authornote{Equal contribution.}
\email{yiyunzhou@zju.edu.cn}
\orcid{0009-0001-5801-8540}
\affiliation{%
  \institution{Zhejiang University}
  \city{Hangzhou}
  \country{China}
}
\author{Zheqi Lv}
\authornotemark[1]
\email{zheqilv@zju.edu.cn}
\orcid{0000-0001-6529-8088}
\affiliation{%
  \institution{Zhejiang University}
  \city{Hangzhou}
  \country{China}
}
\author{Shengyu Zhang}
\email{sy_zhang@zju.edu.cn}
\orcid{0000-0002-0030-8289}
\affiliation{%
  \institution{Zhejiang University}
  \city{Hangzhou}
  \country{China}
}
\author{Jingyuan Chen}
\authornote{Corresponding author.}
\email{jingyuanchen@zju.edu.cn}
\orcid{0000-0003-0415-6937}
\affiliation{%
  \institution{Zhejiang University}
  \city{Hangzhou}
  \country{China}
}

\begin{abstract}
Knowledge Tracing (KT) is a core component of Intelligent Tutoring Systems, modeling learners' knowledge state to predict future performance and provide personalized learning support. Traditional KT models assume that learners' learning abilities remain relatively stable over short periods or change in predictable ways based on prior performance. However, in reality, learners' abilities change irregularly due to factors like cognitive fatigue, motivation, and external stress---a task introduced, which we refer to as Real-time Learning Pattern Adjustment (RLPA). Existing KT models, when faced with RLPA, lack sufficient adaptability, because they fail to timely account for the dynamic nature of different learners' evolving learning patterns. Current strategies for enhancing adaptability rely on retraining, which leads to significant overfitting and high time overhead issues. To address this, we propose Cuff-KT, comprising a controller and a generator. The controller assigns value scores to learners, while the generator generates personalized parameters for selected learners. Cuff-KT controllably adapts to data changes fast and flexibly without fine-tuning. Experiments on five datasets from different subjects demonstrate that Cuff-KT significantly improves the performance of five KT models with different structures under intra- and inter-learner shifts, with an average relative increase in AUC of 10\% and 4\%, respectively, at a negligible time cost, effectively tackling RLPA task. Our code and datasets are 
 fully available at \textcolor{red}{\url{https://github.com/zyy-2001/Cuff-KT}}.
\end{abstract}


\maketitle

\newcommand\kddavailabilityurl{https://doi.org/10.5281/zenodo.15514767}

\ifdefempty{\kddavailabilityurl}{}{
\begingroup\small\noindent\raggedright\textbf{KDD Availability Link:}\\
The source code of this paper has been made publicly available at \url{\kddavailabilityurl}.
\endgroup
}

\section{Introduction}
\label{sec: introduction}
\begin{figure}[htbp]
    \centering
    \includegraphics[width=\linewidth]{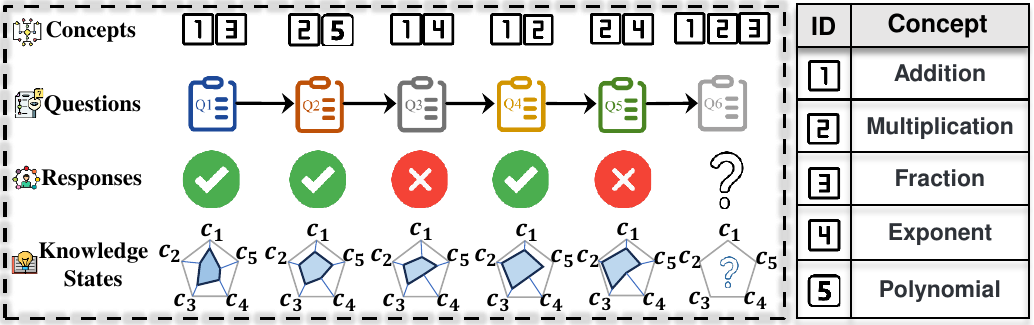}
    \vspace{-0.7cm}
    \caption{Illustration of the Knowledge Tracing (KT).}
    \label{fig: KT}
    \vspace{-0.2cm}
\end{figure}

For nearly a century, researchers have been dedicated to developing Intelligent Tutoring Systems (ITS)~\cite{pressey1926simple, su151612451, zhou2024predictive}. \textit{Knowledge Tracing (KT), as a core component of ITS, aims to model learners' knowledge state during their interactions with ITS to predict their performance on future questions}~\cite{corbett1994knowledge}, as shown in Figure~\ref{fig: KT}. Solving the KT problem can help teachers or systems better identify learners who need further attention and recommend personalized learning materials to them~\cite{liu2021survey, abdelrahman2023knowledge, liu2023simplekt, han2025contrastive, zhou2025revisitingapplicablecomprehensiveknowledge}.

Reviewing the current research on KT~\cite{NIPS2015_bac9162b, zhang2017dynamic, 10.1145/3543507.3583866, 10.1145/3477495.3531939, li2024enhancing, song2022survey, zhou2025disentangled}, we can systematize a dominant paradigm: using learners' historical interaction sequences and encoding them into representations with KT models, and then using these representations to predict future questions or concepts. This paradigm simply assumes that the learner's abilities remain relatively stable in the short term, or that the data changes exhibit certain regularities. That is, previous KT research inherently assumes that learners' abilities remain relatively stable between different short-term stages, or that interaction data from different groups of learners is homogeneous enough to predict the performance of the same learner in the next stage or the different learners in the next group. However, this assumption is difficult to hold in real-world scenarios, as it ignores the dynamic nature of KT. Specifically, the properties of streaming data (\textit{e.g.}, the correct rate of questions or concepts) often change across different stages or groups~\cite{zhang2017dynamic, yang2023generic, 10.1145/3696410.3714743}, indicating that the sequential patterns of learners at different stages or from different groups dynamically vary between historical and future interactions. We refer to these data changes across different stages and groups as intra-learner shift and inter-learner shift, respectively, as defined in Sec.~\ref{sec:task}.

\begin{figure}[t]
    \centering
    \includegraphics[width=0.98\linewidth]{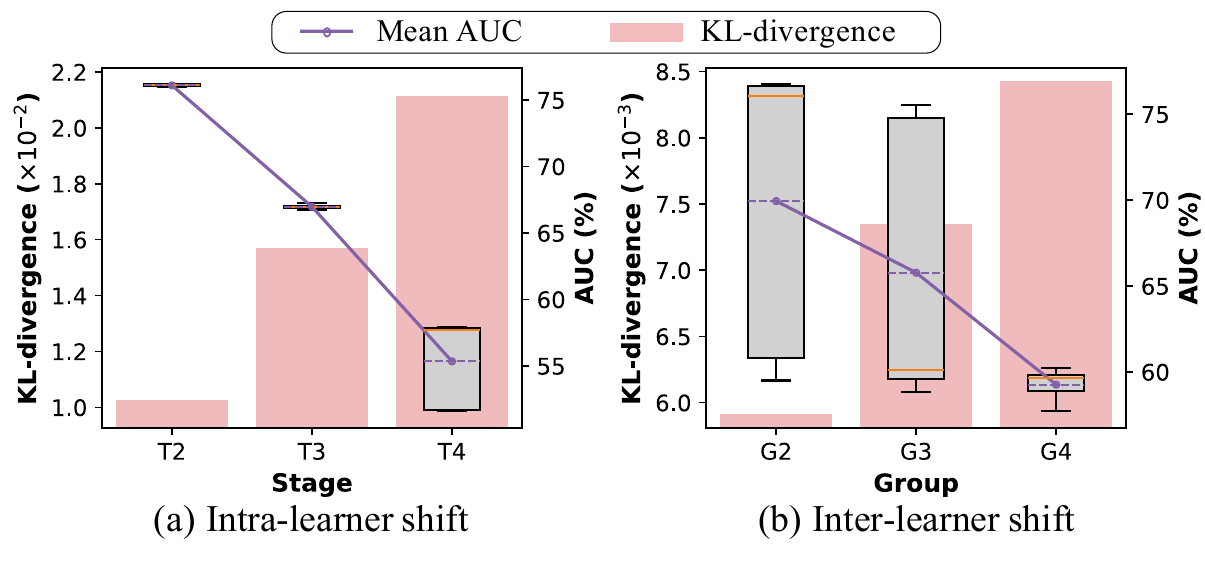}
    \vspace{-0.35cm}
    \caption{Empirical evidence of model generalization deterioration under intra- and inter-learner shifts.}
    \label{fig:problem}
    \vspace{-0.3cm}
\end{figure}

Historical data change caused by varying sequential patterns undermines current KT models, resulting in deteriorated generalization when serving future data. Figure~\ref{fig:problem} provides empirical evidence of this issue. We first divide the assist15~\cite{feng2009addressing} data into 4 non-overlapping parts by stage and group (see Sec.~\ref{seq:implementation} for the division method), and calculate the KL-divergence $w.r.t.$ correct rate between the first part and the other parts. We then train the DKT~\cite{NIPS2015_bac9162b} model on the first part and test it on the remaining parts. Clearly, as the KL-divergence increases across different stages or groups, the model's predictive performance significantly declines. Therefore, it is crucial to enhance the dynamic adaptability of KT models. To this end, we introduce a new task, \textbf{Real-time Learning Pattern Adjustment (RLPA), to address the inability of existing KT models to effectively handle data changes arising from differing learning patterns across various stages or groups.}

To tackle RLPA, a well-known generalization technique is to retrain (\textit{e.g.}, fine-tune) the pre-trained KT model based on data from the recent stage or group to achieve personalized learning~\cite{houlsby2019parameter, zaken2021bitfit, han2024parameter}. 
Although fine-tuning-based approaches are promising, they may not be the optimal solution due to two key challenges:
\text{(i)} \textbf{Overfitting}. To achieve personalized learning, fine-tuning-based approaches often require retraining the model based on very limited samples with rapidly changing data, which may lead to overfitting, potentially reducing its ability to generalize to future data~\cite{hawkins2004problem, kuhn2013over, zhou2025colacollaborativelowrankadaptation}.
\text{(ii)} \textbf{High time overhead}. Fine-tuning is very time-consuming as it requires extensive gradient computations to update model parameters, which is cumbersome in real-world scenarios where real-time requirements are common~\cite{xu2023parameter, xia2024understanding}.
Therefore, fine-tuning-based methods must carefully balance the need to adapt to recent data and maintain robustness to achieve generalization. These challenges prompt us to reconsider the design of better solutions to the RLPA.

\begin{sloppypar}
Towards this end, we propose a novel method to trackle RLPA in KT, called \colorbox{mygray}{\underline{\textbf{C}}ontrollable}, \colorbox{mygray}{t\underline{\textbf{U}}ning-free}, \colorbox{mygray}{\underline{\textbf{F}}ast}, and \colorbox{mygray}{\underline{\textbf{F}}lexible} \underline{\textbf{K}}nowledge \underline{\textbf{T}}racing (\textbf{Cuff-KT}). Unlike fine-tuning-based approaches that produce \textbf{updated} parameters, the core idea of Cuff-KT is to learn a model parameter generator specific to the current stage or group, generating \textbf{updating} personalized parameters for valuable learners (\textit{e.g.}, those showing significant progress or regression), achieving adaptive generalization. Our Cuff-KT consists of two modules: a controller and a generator. \textbf{When the data of learners changes due to varying learning patterns, the KT model generalizes worse to the current data and tends to make incorrect evaluations. This implies that the benefit of generating parameters is significant, as generated parameters can appropriately model the current data distribution.} The controller, while considering the fine-grained distance between knowledge state distributions across various  concepts, is also inspired by the Dynamic Assessment Theory~\cite{vygotsky1978mind, carlson1992principles, anton2021dynamic} and integrates coarse-grained changes in correct rates, assigning a value score to each learner.
The generator generates parameters for learners selected based on the assigned value scores\footnote{The larger a learner's value score, the more likely they are to be selected.} by the controller and enhances adaptability. Specifically, considering the relative relationship between question difficulty and learner ability~\cite{rasch1993probabilistic, 10.1145/3477495.3531939} and inspired by the dual-tower model in recommendation systems~\cite{huang2013learning, covington2016deep}, the generator models questions and responses separately, extracts features through a sequential feature extractor, simulates the distribution of real-time samples from the current stage or group to achieve adaptive generalization through our designed state-adaptive attention, and finally reduces the parameter size through low-rank decomposition. Notably, our generator can be inserted into into any layer or generate parameters for any layer.
\end{sloppypar}
Our contributions are summarized as follows:

\begin{itemize}[leftmargin=*]
    \item{We introduce a new task, \textbf{RLPA}, which enhances the adaptability of existing KT models in the realm of personalized learning, addressing the challenges arising from data changes caused by varying sequential patterns of learners across different stages or groups.}
    \item{We propose \textbf{Cuff-KT}, a controllable, tuning-free, fast, and flexible general neural KT method, which can effectively generate parameters aligned with the current stage or group's learner data regularities and insert them into any layer of existing KT models. It is noteworthy that Cuff-KT is model-agnostic.}
    \item{We instantiate five representative KT models with different structures. Experiments on five datasets from different subjects demonstrate that our proposed Cuff-KT generally improves current KT models under intra- and inter-learner shifts. Specifically, the AUC metric, which is the most commonly used in KT, has relatively increased by 10\% and 4\% on average under intra- and inter-learner shifts, respectively, proving that Cuff-KT can effectively tackle RLPA in KT.}
\end{itemize}

\section{Related Work}
\label{sec:related_work}

\begin{sloppypar}
Knowledge tracing (KT), the task of dynamically modeling a learner's knowledge state over time, traces its origins back to the early 1990s, with early notable contributions by Corbett and  Anderson~\cite{corbett1994knowledge}. However, with the rise of deep learning, KT research has gained significant momentum, leading to the development of more sophisticated and refined models capable of capturing the intricate dynamics of learner behavior~\cite{NIPS2015_bac9162b, 10.1145/3231644.3231647, 10.1145/3477495.3531939, 10.1145/3543507.3583866}. \textcolor[rgb]{1,0,0}{DKT}~\cite{NIPS2015_bac9162b} first applies \textcolor[rgb]{1,0,0}{LSTM} to KT to model the complex learners' cognitive process, bringing a leap in performance compared to previous KT models (\textit{e.g.}, BKT~\cite{corbett1994knowledge}). Subsequently, various neural architectures (especially attention mechanism) begin to be introduced into KT~\cite{pandey2019self, nakagawa2019graph, 10.1145/3394486.3403282, 10.1145/3340531.3411994}. \textcolor[rgb]{1,0.8,0}{DKVMN}~\cite{zhang2017dynamic}, based on Memory-Augmented Neural Networks (\textcolor[rgb]{1,0.8,0}{MANN}), dynamically stores students' mastery of concepts using the key matrix and the value matrix. \textcolor{blue}{AT-DKT}~\cite{10.1145/3543507.3583866} addresses the issues of sparse representation and personalization in DKT (\textcolor{blue}{LSTM}) by introducing two auxiliary learning tasks: question tagging prediction and individualized prior knowledge prediction, combined with an \textcolor{blue}{attention} module. Meanwhile, \textcolor[rgb]{0.0,0.9,0.5}{stableKT}~\cite{li2024enhancing} uses a multi-head aggregation module combining dot-product and hyperbolic \textcolor[rgb]{0.0,0.9,0.5}{attention} to achieve length generalization. Additionally, incorporating learning-related information with \textcolor[rgb]{0.54,0.17,0.89}{customized neural networks} has been explored. For instance, \textcolor[rgb]{0.54,0.17,0.89}{DIMKT}~\cite{10.1145/3477495.3531939} improves KT performance by establishing relationship between learners' knowledge states and question difficulty levels with a customized adaptive sequential neural network. Recently, some studies on the applicability and comprehensiveness of KT have gained attention~\cite{shen2024survey, zhou2025revisitingapplicablecomprehensiveknowledge}. KT models like DKVMN and stableKT face challenges in scalability due to their poor applicability, which can be addressed by transforming the input and output formats.
\end{sloppypar}

However, surprisingly, to our knowledge, there is a lack of attention to adaptability in KT research, which severely affects the generalization of KT models across different distributions. Thanks to the well-known fine-tuning-based methods, the adaptability in KT has been enhanced to some extent. However, the challenges posed by overfitting and high time overhead of fine-tuning-based methods make it difficult to be effectively applied in real-world scenarios~\cite{lv2023duet}. Even the recently proposed parameter-efficient fine-tuning-based methods (\textit{e.g.}, Adapter-based tuning ~\cite{houlsby2019parameter} and Bias-term Fine-tuning~\cite{zaken2021bitfit}) still incur non-negligible time cost and cannot avoid the potential risk of overfitting. Our Cuff-KT, in contrast, updates KT models under dynamic distributions through parameter generation, eliminating the need for retraining and providing a new perspective on enhancing adaptability in the KT community.

\section{Methodology}
\label{sec:method}

\setlength{\abovedisplayskip}{3pt}
\setlength{\belowdisplayskip}{3pt}

In this section, we first define the problem of KT and formalize the RLPA task in KT, then introduce our proposed Cuff-KT method.

\subsection{Problem Formulations}
\label{sec: formulations}

\subsubsection{\textbf{Knowledge Tracing}}
\label{sec:definion}

Formally, let $\mathcal{S}$, $\mathcal{Q}$, and $\mathcal{C}$ denote the sets of learners, questions, and concepts, respectively. For each learner $s\in \mathcal{S}$, their interactions are represented  $X^s=\{x\}_{i=1}^{k}$ at time-step $k$, where the interaction $x$ is defined as a $4$-tuple, \textit{i.e.}, $x=(q,\ \{c\},\ r,t)$, where $q\in \mathcal{Q}, \{c\} \subset \mathcal{C}, r, t$ represent the question attempted by the learner $s$, the concepts associated with the question $q$, the binary variable indicating whether the learner responds to the question correctly ($1$ for correct, $0$ for incorrect), and the timestamp of the learner's response respectively. The goal of KT is to predict $\hat{r}_{k+1}$ given the learner's historical interactions $X^s$ and the current question $q_{k+1}$ at time-step $k+1$.

\subsubsection{\textbf{RLPA Task}}
\label{sec:task}
RLPA aims to address two common shift issues (intra- and inter- shifts) in KT to enhance the adaptability of existing models. An interaction sequence of a learner $s$ can be divided into multiple stages, assuming each stage has a length of $L$. At time-step $u$, the representation of the learner's interaction in that stage is $X_u^s=X^s_{u:u+L-1}$. Intra-learner shift is defined as: for any time-step $u\neq v$,
\begin{align}
    \lvert d(\chi^s_u, \chi^s_v) \rvert > \delta,\label{eq: intra-shift}
\end{align}
where $\delta$ is a small threshold. $\chi^s_u$ and $\chi ^s_v$ represent the distributions of $X^s_u$ and $X^s_v$ respectively. $d$ is a distance function (\textit{e.g.}, KL divergence). Similarly, inter-learner shift is:
\begin{align}
    \lvert d(\chi^s_u, \chi^{s^{\ast}}_u) \rvert > \delta,\label{eq: inter-shift}
\end{align}
where $\chi^s_u$ and $\chi^{s^{\ast}}_u$ represent the distributions of learners $s$ and $s^{\ast}$ at time-step $u$, respectively.

When equations~(\ref{eq: intra-shift}) or~(\ref{eq: inter-shift}) hold, the goal of RLPA is to adjust the parameters of the existing KT model in real-time so that the predicted distribution $\hat{\chi}^s_v$ or $\hat{\chi}^{s^{\ast}}_u$ is as close as possible to the actual distribution:
\begin{align}
    \min_{\hat{\chi}^s_v}\sum_{x}\chi^s_v(x)\text{log}(\frac{\chi^s_v(x)}{\hat{\chi}^s_v(x)}) \ \text{or} \ \min_{\hat{\chi}^{s^{\ast}}_u}\sum_{x}\chi^{s^{\ast}}_u(x)\text{log}(\frac{\chi^{s^{\ast}}_u(x)}{\hat{\chi}^{s^{\ast}}_u(x)}),
\end{align}
where $x$ is a variable in the sample space.

\vspace{-0.1cm}
\subsection{Cuff-KT}
\vspace{-0.1cm}

\begin{figure*}[t]
    \centering
    \includegraphics[width=\linewidth]{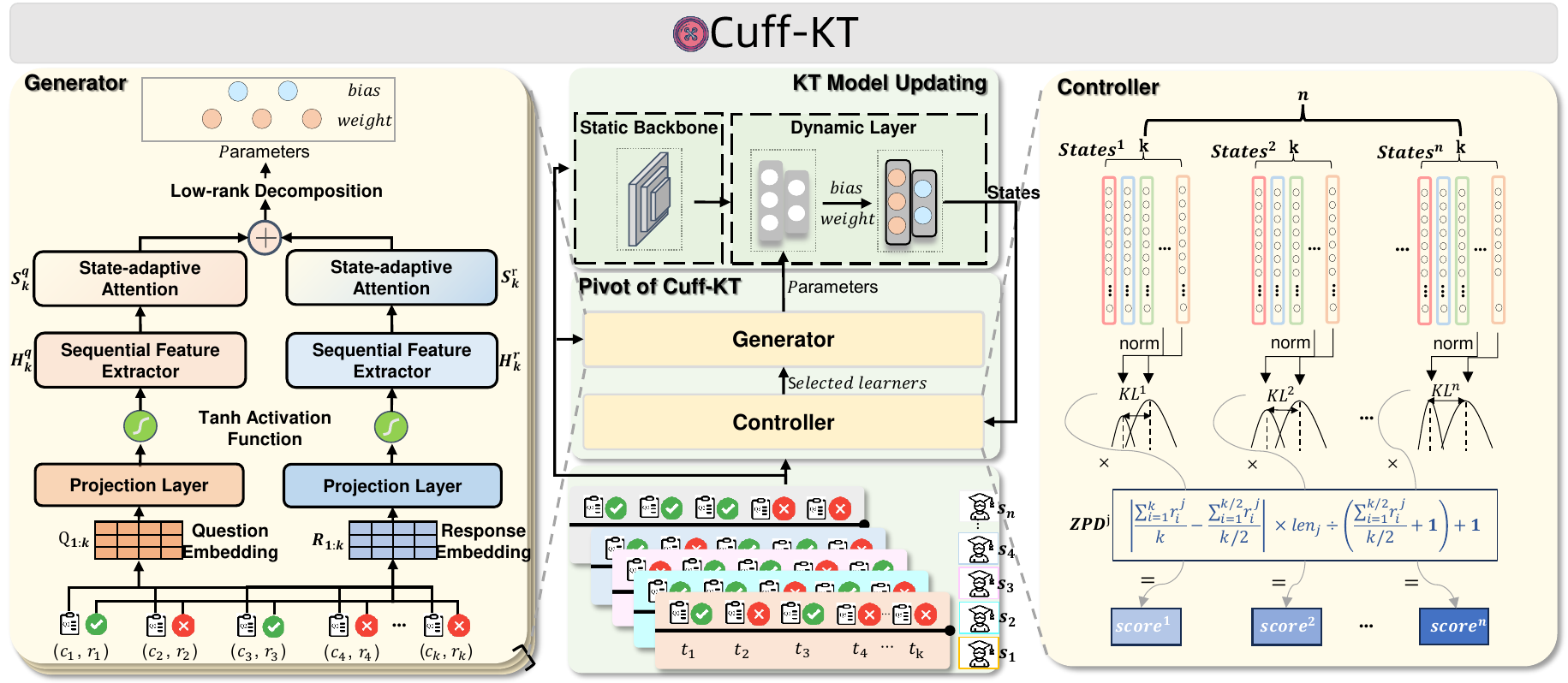}
    \vspace{-0.6cm}
    \caption{Overview of proposed Cuff-KT method. Cuff-KT selects learners with parameter update value for the generator through the controller. The generator produces adaptive parameters for the decomposed dynamic layers (\textit{e.g.}, output layer) in the KT model based on real-time data to update the model.}
    \vspace{-0.3cm}
    \label{fig: method}
\end{figure*}

Figure~\ref{fig: method} illustrates an overview of our Cuff-KT method, which consists of two modules: (a) \textbf{Controller} identifies learners with valuable parameter update potential, aiming to \textbf{reduce the cost of parameter generation}. (b) \textbf{Generator} adjusts network parameters for existing KT models at different stages or for different groups, aiming to \textbf{enhance adaptive generalization}. In our setup, the KT model is decoupled into a static backbone and a dynamic layer. The generator can be inserted into any layer of the KT model or generate parameters for any layer (dynamic layer). Finally, we introduce the training strategy for Cuff-KT.

\subsubsection{\textbf{Controller}}
\label{sec: controller}

The controller can identify learners with dramatic changes in their knowledge state distribution (\textit{i.e.}, valuable learners, often showing progress or regression), aiming to reduce the cost of parameter generation. The controller comprehensively considers both fine-grained and coarse-grained changes in the knowledge states of different learners, as described below.

\textbf{Fine-grained Changes}.
At time-step $k$, the KT model models knowledge states $\text{States}^j$ (\textit{i.e.}, proficiency scores ranging from $0$ to $1$ for $\left | \mathcal{C} \right |$ concepts over $k$ time steps) for learner $s^j$ with number $j$ ($1 \leq j \leq n$, where $n$ is the total number of learners) at different time steps. The $\text{States}^j$ is utilized by the controller to measure the fine-grained distance (\textit{e.g.}, KL-divergence) between the knowledge state distributions across various concept at the intermediate time-step $k/2$ and current time-step $k$ (
\textbf{the knowledge state predicted at time-step $k/2$ is trustworthy, and there is a gap between it and the knowledge state at the time-step $k$  to facilitate differentiation}):
\begin{flalign}
    &\left\{
    \begin{aligned}
        &{\text{States}^\ast}_{k/2}^j = \text{norm}(\text{States}_{k/2}^j),\\
        &{\text{States}^\ast}_k^j = \text{norm}(\text{States}_k^j),\\
        &\text{KL}^j = \sum_{c\in \mathcal{C}} {\text{States}^\ast}_k^j(c)\text{log}\frac{{\text{States}^\ast}_k^j(c)}{{\text{States}^\ast}_{k/2}^j(c)} + 1,
    \end{aligned}
    \right.
\end{flalign}
where $\text{norm}(\cdot)$ denotes the normalization operation.

\textbf{Coarse-grained Changes}.
However, focusing solely on fine-grained changes might not capture the overall knowledge state changes of the learners. Focusing on coarse-grained changes is meaningful. For example, \textbf{when underperforming learners start to achieve good grades, this contradicts the prior assumptions of KT models and should be considered more noteworthy than the progress of high-performing learners, and vice versa.} Its effectiveness is verified in Appendix~\ref{sec: ablation_controller}. The Zone of Proximal Development (ZPD) is a core concept in Dynamic Assessment Theory~\cite{vygotsky1978mind}. It refers to the gap between a learner's current independent ability level and the potential level that could be reached with the help of other mediums (\textit{e.g.}, ITS). It describes the overall changes in the learner's knowledge state (\textit{i.e.}, progress or regression). Inspired by this, we consider the overall correct rate at the intermediate time-step $k/2$ as the lower limit of the ZPD, and the correct rate at the current time-step $k$ as the upper limit or near-upper limit of the ZPD. We use the rate of change as a quantitative indicator of the $\text{ZPD}^j$ of learner $s^j$:
\begin{flalign}
&\text{ZPD}^j=\left|\frac{\sum_{i=1}^k r_i^j}{k}-\frac{\sum_{i=1}^{k / 2} r_i^j}{k / 2}\right| \times \operatorname{len}_j \div\left(\frac{\sum_{i=1}^{k / 2} r_i^j}{k / 2} + 1\right ) + 1, \label{eq:zpd}
\end{flalign}
where $\operatorname{len}_j$ is the actual length of questions attempted by learner $s^j$ ($\operatorname{len}_j \leq k$, when $\operatorname{len}_j < k$, the missing sequence, \textit{e.g.}, concepts sequence, is often padded with $0$), which reflects the reliability of the $\text{ZPD}^j$, with a larger $\operatorname{len}_j$ indicating more reliable results. The equation~(\ref{eq:zpd}) reliably reflects the overall degree of change in the progress of  high-performing learners or the regression of underperforming learners (\textit{e.g.}, \textbf{an underperforming learner $s^j$ initially shows a smaller value of $\sum_{i=1}^{k/2}r_i$, and as progress is reflected in the increase of $\sum_{i=1}^{k}r_i$, the $\text{ZPD}^j$ increases accordingly, making it more likely for him/her to receive attention}).

Finally, the controller assigns a value score to learner $s^j$:
\begin{flalign}
    \text{score}^j = \text{KL}^j\times \text{ZPD}^j.
\end{flalign}
It can be observed that $\text{KL}^j$ and $\text{ZPD}^j$ are positive, which avoids any absolute impact on the $\text{score}^j$ when either one is 0. Notably,the controller can identify learners who have shown significant progress or regression, which is beneficial for teachers or ITS to pay further attention to them.

\subsubsection{\textbf{Generator}}
\label{sec: generator}

The generator can generate personalized dynamic parameters for learners determined by the controller based on real-time samples from different stages or groups, aiming to improve the adaptive generalization for continuously changing distributions. We first introduce the generator's feature extraction, then propose our designed state-adaptive attention, and finally discuss generating parameters through low-rank decomposition. For convenience, we have omitted the superscript of the learner numbers.

\textbf{Sequential Feature Extractor (SFE)}. 
At time-step $k$, the SFE takes $\{(c_i, r_i)\}_{i=1}^k$ as input (\textbf{KT often substitutes the question $q_i$ with the concept $c_i$ to address the sparsity issue}), considering the relative relationship between question difficulty and learner ability~\cite{rasch1993probabilistic} and inspired by the dual-tower model in recommendation systems~\cite{huang2013learning, covington2016deep}, embedding the questions $c_{1:k}$ and responses $r_{1:k}$ into vector spaces $Q_{1:k}\in \mathbb{R}^d$ and $R_{1:k}\in \mathbb{R}^d$, respectively (\textbf{$d$ is the dimension of the embedding}). After non-linearization, features $H_k^q\in \mathbb{R}^{d_{in}}$ and $H_k^r\in \mathbb{R}^{d_{in}}$ (\textbf{$d_{in}$ is the input dimension of the dynamic layer}) are extracted through the SFE (\textit{e.g.}, GRU):
\begin{flalign}
    &\left\{
    \begin{aligned}
        &H_k^q = \text{SFE}(\text{Tanh}(Q_{1:k}W_1 + b_1)),\\
        &H_k^r = \text{SFE}(\text{Tanh}(R_{1:k}W_2 + b_2)),
    \end{aligned}
    \right.
\end{flalign}
where $W_1\in \mathbb{R}^{d_{in}\times d}$, $W_2\in \mathbb{R}^{d_{in}\times d}$, $b_1\in \mathbb{R}^{d_{in}}$, $b_2\in \mathbb{R}^{d_{in}}$ are learnable parameters in the projection layer. $\text{Tanh}(\cdot)$ is the activation function.

\textbf{State-adaptive Attention (SAA)}.
SAA is responsible for adaptive generalization of the extracted question and response features, considering both change in concept correct rate (\textit{i.e.}, difficulty) and the time of the change in knowledge state. Intuitively, \textbf{the greater the change in difficulty, indicating more significant progress or regression, and the longer the time since the last response, the more likely a sudden change in knowledge state can occur. These time-steps should receive more attention}. Therefore, the definition of SAA is as follows:
\begin{flalign}
    &\left\{
    \begin{aligned}
        &\text{SAA}(X_k) = \text{Concat}(\text{head}_1, \cdots, \text{head}_h)W_h,\\
        &\text{head}_i=\text{Attention}^\ast(Q=X_k^{/h},K=X_k^{/h},V=X_k^{/h}),\\
        &\text{Attention}^\ast(Q,K,V)=\text{softmax}^\ast(X=\frac{QK^T}{\sqrt{d/h}})V,\\
        &\text{softmax}^\ast(X)=\text{attn}_w(c_{1:k}, r_{1:k}, t_{1:k})\cdot \text{softmax}(X),\\
        &\text{attn}_w(c_{1:k}, r_{1:k}, t_{1:k}) = \text{dist}_d(c_{1:k}, r_{1:k})\cdot \text{dist}_t(c_{1:k}, t_{1:k}),
    \end{aligned}
    \right.
\end{flalign}
where $h$ is the number of attention heads and $W_h \in \mathbb{R}^{d_{in}\times d_{in}}$. $X_k^{/h}$ represents splitting the $d_{in}$ dimensions of $X_k$ into $h$ parts. $\text{dist}_d$ and $\text{dist}_t$ represent the changes in difficulty and time, respectively. $t_{1:k}$ is the sequence of timestamps up to time-step $k$, as shown in Sec.~\ref{sec:definion}. At time-step $i \in [1, k]$, $\text{dist}_d(c_i, r_i)$ and $\text{dist}_t(c_i, t_i)$ are respectively:
\begin{equation}
\begin{split}
    \begin{cases}
       1, & \text{if } i=1,\\
       \left(\frac{\sum_{j=1}^{i}r_j[c_j = c_i]}{\sum_{j=1}^i 1[c_j = c_i]} - \frac{\sum_{j=1}^{i-1}r_j[c_j = c_i]}{\sum_{j=1}^{i-1} 1[c_j = c_i]}\right) + 1. & \text{else}
    \end{cases}
\end{split}\label{eq:dist_d}
\end{equation}
\begin{equation}
\begin{split}
    \begin{cases}
       1, & \text{if } j = \text{max}\{ k \mid k < i \text{ and } c_k = c_i \} = \emptyset,\\
       \frac{t_i - t_j}{t_i - t_1}. & \text{else}
    \end{cases}
\end{split}\label{eq:dist_t}
\end{equation}

Equations~(\ref{eq:dist_d}) and~(\ref{eq:dist_t}) measure the fine-grained variations in question difficulty and the time interval between responses to the same question, respectively. Finally, the representations $S_k^q$ and $S_k^r$ of question difficulty and learner ability are obtained by SAA:
\begin{flalign}
    S_k^q = \text{SAA}(H_k^q), S_k^r = \text{SAA}(H_k^r),
\end{flalign}
where \textbf{$S_k^q$ and $S_k^r$ characterize the difficulty distribution of questions and the ability distribution of learners, respectively, based on real-time data from the current stage or group.} SAA is the core component of the generator, and we will further discuss its importance in Sec.~\ref{sec: ablation}.

\textbf{Low-rank Decomposition}.
Before performing low-rank decomposition on the parameters, the learned question difficulty $S_k^q$ and learner ability $S_k^r$ are uniformly expressed as the generalized information feature $S_k$ that characterizes the interaction distribution of learners:
\begin{flalign}
    S_k = S_k^q + S_k^r,
\end{flalign}
Finally, parameters (\textit{i.e.}, weight and bias) are generated through $S_k$ for the dynamic layer:
\begin{flalign}
    &\left\{
    \begin{aligned}
        &\text{weight} = S_kW_w + b_w,\\
        &\text{bias} = S_kW_b + b_b,
    \end{aligned}
    \right.
\end{flalign}
where $W_w\in \mathbb{R}^{(d_{in}\times d_{out})\times d_{in}}$, $b_w\in \mathbb{R}^{d_{in}\times d_{out}}$, $W_b\in \mathbb{R}^{d_{out}\times d_{in}}$, $b_b\in \mathbb{R}^{d_{out}}$ are learnable parameters. $d_{out}$ is the output dimension of the dynamic layer.
However, it can be observed that the parameter size of $W_w$ is too large, which increases computational resources and the risk of overfitting. Inspired by LoRA~\cite{hu2021lora}, $W_w$ is decomposed into low-rank matrices to obtain the final weight:
\begin{flalign}
    \text{weight} = S_kW_{w_1}W_{w_2} + b_w,
\end{flalign}
where $W_{w_1}\in \mathbb{R}^{\text{rank}\times d_{in}}$, $W_{w_2}\in \mathbb{R}^{(d_{in}\times d_{out})\times \text{rank}}$ are learnable parameters, and $\text{rank}\ll d_{in}$ is a very small value (\textit{e.g.}, $1$). In Sec.~\ref{sec: ablation}, we will further analyze the effects of different $\text{ranks}$.

It's noted that the generator can generate parameters for the dynamic layer, given the input dimension $d_{in}$ and output dimension $d_{out}$. In our experiments, the dynamic layer defaults to the output layer of the KT model. Meanwhile, \textbf{for KT models that are not applicable, as discussed in Sec.~\ref{sec:related_work}, we refer to~\cite{zhou2025revisitingapplicablecomprehensiveknowledge} to make them compatible with our Cuff-KT in order to avoid label leakage.}

\subsubsection{\textbf{Model Training}}
\label{sec: training}
All learnable parameters are trained by minimizing the binary cross-entropy between $r_i$ and $\hat{r}_i$, \textit{i.e.},
\begin{flalign}
    \mathcal{L} = -\sum_{i=2}^{k+1}{r_{i}\text{log}({\hat{r}}_{i})+(1-r_{i})\text{log}(1-{\hat{r}}_{i})}.
\end{flalign}

\section{Experiments}
\label{sec:experiments}

In this section, we demonstrate the superiority of our proposed Cuff-KT and the impact of its different components through experiments. Specifically, the experimental evaluation is divided into
\texttt{(i)} the controllability of parameter generation (Sec.~\ref{sec: controllability}), 
\texttt{(ii)} prediction accuracy, quantifying the effectiveness of tackling RLPA (Sec.~\ref{sec: prediction}), 
\texttt{(iii)} the flexible application of Cuff-KT (Sec.~\ref{sec: cuff-kt+}), and 
\texttt{(iv)} the impact of  dual-tower modeling, SFE, SAA, and low-rank decomposition in Cuff-KT (Sec.~\ref{sec: ablation}).

\subsection{Experimental Setup}

\subsubsection{\textbf{Datasets}}

We conduct extensive experiments on five benchmark datasets from different subjects over the past decade: assist15~\cite{feng2009addressing}, assist17, comp~\cite{hu2023ptadisc}, xes3g5m~\cite{liu2024xes3g5m}, and dbe-kt22~\cite{abdelrahman2022dbe}. The introduction, preprocessing details, and statistics of these datasets can be found in Appendix~\ref{sec: datasets}.

\subsubsection{\textbf{Baselines}}

\begin{sloppypar}
We select five representative KT models (\textcolor[rgb]{1,0,0}{DKT}~\cite{NIPS2015_bac9162b}, \textcolor[rgb]{1,0.8,0}{DKVMN}~\cite{zhang2017dynamic}, \textcolor{blue}{AT-DKT}~\cite{10.1145/3543507.3583866}, \textcolor[rgb]{0.0,0.9,0.5}{stableKT}~\cite{li2024enhancing} and \textcolor[rgb]{0.54,0.17,0.89}{DIMKT}~\cite{10.1145/3477495.3531939}) with different structures as the backbone models for optimization. For the introduction of these models, please refer to Appendix~\ref{sec: baselines}. It's noted that, as categorized in ~\cite{zhou2025revisitingapplicablecomprehensiveknowledge} regarding the input and output formats of KT models, DKVMN and stableKT are inapplicable models. To prevent label leakage when directly applying them to Cuff-KT, we convert them into applicable KT models according to ~\cite{zhou2025revisitingapplicablecomprehensiveknowledge} to ensure compatibility with our Cuff-KT. The performance loss caused by the conversion can be referred to in ~\cite{zhou2025revisitingapplicablecomprehensiveknowledge} and is not within the scope of this paper. Instead, we focus more on the further optimization of different methods.
\end{sloppypar}

We compare Cuff-KT with these five backbone models and three classic fine-tuning-based methods  (\textbf{The LoRA~\cite{hu2021lora} method is not chosen as a baseline because, in our experiments, the limited number of fine-tuning samples leads to severe overfitting for LoRA, which significantly degrades model performance, as stated in the Sec.~\ref{sec: introduction}}): Full Fine-tuning (FFT), Adapter-based tuning (Adapter)~\cite{houlsby2019parameter} and Bias-term Fine-tuning (BitFit)~\cite{zaken2021bitfit}. The introduction of these fine-tuning-based methods can be found in Appendix~\ref{sec: baselines}.

\begin{figure*}[t]
    \centering
    \includegraphics[width=\linewidth]{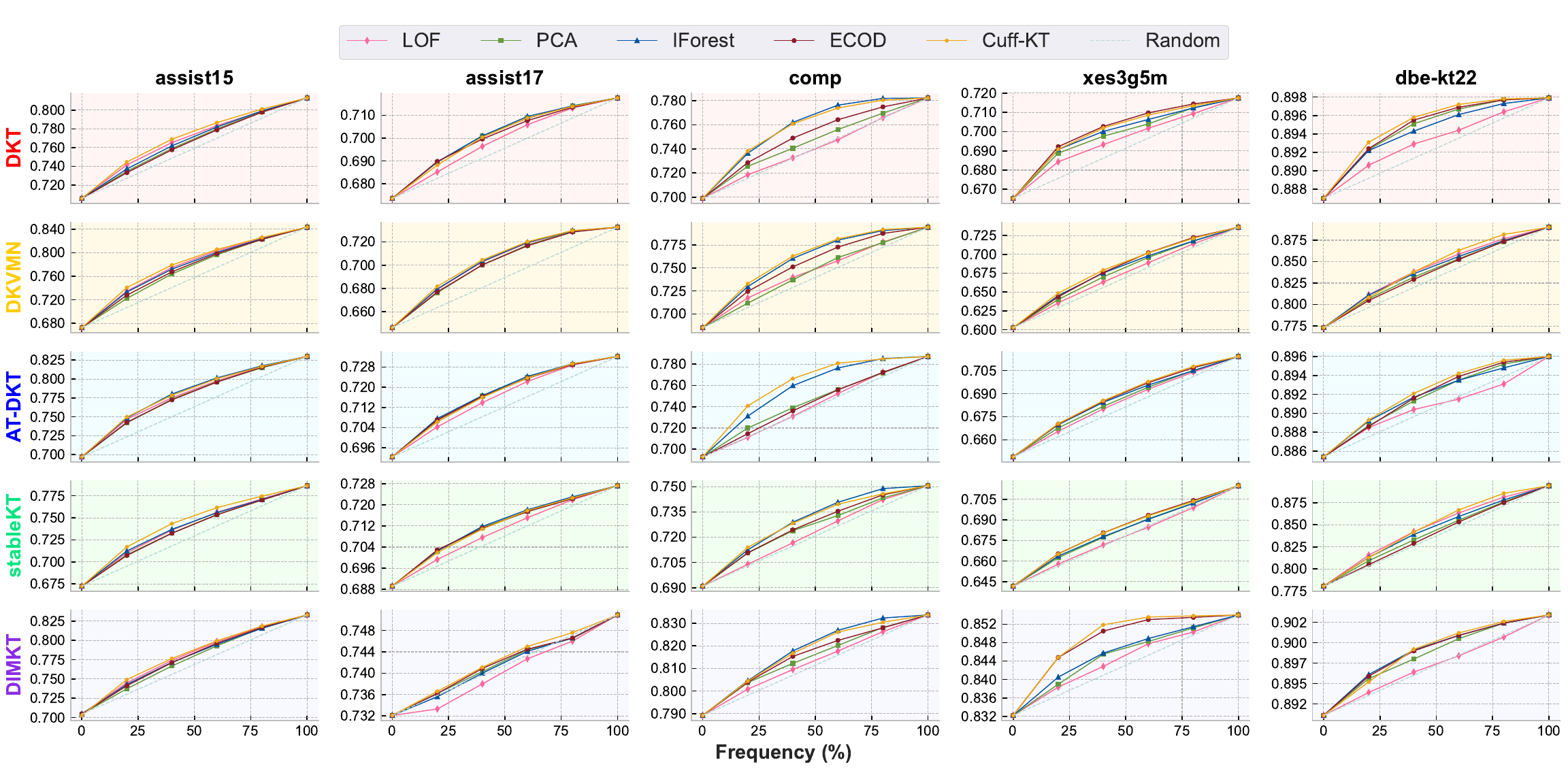}
    \vspace{-0.8cm}
    \caption{Performance comparison of Cuff-KT and four unsupervised anomaly detection algorithms at different frequencies.}
    \vspace{-0.3cm}
    \label{fig: control_dkt}
\end{figure*}

\subsubsection{\textbf{Implementation}}
\label{seq:implementation}

\begin{sloppypar}
We implement all models using Pytorch~\cite{paszke2019pytorch} on a Linux server with two GeForce RTX 3090s and an NVIDIA A800-SXM4 (80G) GPU (used when the model cannot fit into the memory of the RTX 3090s). We used the Adam~\cite{kingma2014adam} optimizer with a learning rate of 0.001, and a batch size of 512. The embedding dimension for all models is fixed at 32. The rank of the generator in Cuff-KT is set to 1 by default. We split the historical interactions into training, validation, fine-tuning, and test datasets (7:1:1:1) based on timestamps and groups, respectively. Specifically, \textbf{for learners at different stages, we divide them according to the timestamps of when they started responding to questions. For learners in different groups, we sort them based on the changes in their knowledge states between two key timestamps. That is, we use DKT to encode each learner’s interaction history and use the distance (\textit{e.g.}, KL divergence) between the prediction distributions for each concept at the intermediate and current timestamps as the basis for division}. An early stopping strategy is applied if the AUC on the validation set does not increase for 10 epochs. The experiments are repeated 5 times under random seeds 0 to 4 and the average performance is reported. Following the previous works~\cite{NIPS2015_bac9162b, 10.1145/3477495.3531939, 10.1145/3543507.3583866}, the evaluation metrics include Area Under the ROC Curve (AUC) and Root Mean Square Error (RMSE). Moreover, in Sec.~\ref{sec: prediction}, we evaluate the additional Time Overhead (TO) of all methods based on the backbone.
\end{sloppypar}

\subsection{\colorbox{mygray}{\underline{C}ontrollable} Parameter Generation}
\label{sec: controllability}

According to~\cite{lv2023ideal, lv2024intelligent, lv2024semantic}, data with drastic distribution updates often bring benefits to the model (\textit{e.g.}, learner data with significant progress or regression contributes more to model generalization than mediocre learner data). Previous studies ~\cite{patcha2007overview, chandola2009anomaly, jiang2023anomaly, cao2023anomaly} have shown that anomaly detection algorithms can be used to detect distribution changes over time. So we select four representative unsupervised anomaly detection algorithms from pyod library~\cite{zhao2019pyod} as comparison baselines for the controller in Cuff-KT: LOF~\cite{breunig2000lof}, PCA~\cite{shyu2003novel}, IForest~\cite{4781136}, and ECOD~\cite{li2022ecod}. Detailed descriptions of these four algorithms are shown in Appendix~\ref{sec: algorithms}.

Since intra-learner shift is more sensitive to data distribution and more common in real-world scenarios, we conduct experiments under this setting and use AUC as the evaluation metric. Figure~\ref{fig: control_dkt} shows the performance results under intra-learner shift when the controller selects learners with different frequencies for the generator. We can see that anomaly detection algorithms (especially IForest and ECOD) consistently outperform the random selection, demonstrating the effectiveness of using anomaly detection algorithms to detect distribution changes. Moreover, our Cuff-KT generally performs better than these algorithms, indicating that Cuff-KT is more capable of identifying learners whose model generalization deteriorates due to distribution changes. We attribute Cuff-KT's breakthrough to the Dynamic Assessment Theory~\cite{vygotsky1978mind}, and the ablation study of different components of the controller in Cuff-KT is shown in Appendix~\ref{sec: ablation_controller}.

\subsection{\colorbox{mygray}{T\underline{u}ning-Free} and \colorbox{mygray}{\underline{F}ast} Prediction}
\label{sec: prediction}

\begin{table*}[h]
\setlength{\tabcolsep}{1mm}
\centering
\caption{Performance comparison between different methods under intra-learner shift. $^\ast$ and $^{\ast\ast}$ indicate that the improvements over the strongest baseline are statistically significant, with \textit{p} \textless 0.05 and \textit{p} \textless 0.01, respectively. $^{\dagger}$ indicates that the model is not directly applicable, but can be made compatible with Cuff-KT (potentially leading to performance loss); for details, please refer to~\cite{zhou2025revisitingapplicablecomprehensiveknowledge}. It's noted that not all backbones are optimal, and the improvements across different methods are noteworthy.}
\vspace{-0.4cm}
\label{tab: result_intra-learner}
\resizebox{\textwidth}{!}{%
\begin{tabular}{@{}cc|c|c|c|c|c|c|c|c|c|c|c|c|c|c@{}}
\toprule[1.5pt]
\multicolumn{1}{c|}{\textbf{Dataset}}                                           & \multicolumn{3}{c|}{\textbf{assist15}} & \multicolumn{3}{c|}{\textbf{assist17}} & \multicolumn{3}{c|}{\textbf{comp}}     & \multicolumn{3}{c|}{\textbf{xes3g5m}}  & \multicolumn{3}{c}{\textbf{dbe-kt22}} \\ \cmidrule(l){1-1} \cmidrule(l){2-4} \cmidrule(l){5-7} \cmidrule(l){8-10} \cmidrule(l){11-13} \cmidrule(l){14-16} 
\multicolumn{1}{c|}{\textbf{Metric}} & \textbf{AUC $\uparrow$}    & \textbf{RMSE $\downarrow$}   & \textbf{TO (ms) $\downarrow$} & \textbf{AUC $\uparrow$}    & \textbf{RMSE $\downarrow$}   & \textbf{TO (ms) $\downarrow$} & \textbf{AUC $\uparrow$}    & \textbf{RMSE $\downarrow$}   & \textbf{TO (ms) $\downarrow$} & \textbf{AUC $\uparrow$}    & \textbf{RMSE $\downarrow$}   & \textbf{TO (ms) $\downarrow$} & \textbf{AUC $\uparrow$}    & \textbf{RMSE $\downarrow$}   & \textbf{TO (ms) $\downarrow$} \\ \cmidrule(l){1-16} 
\rowcolor[HTML]{F8F8F8}\multicolumn{1}{l|}{\cellcolor{white}\textcolor[rgb]{1,0,0}{DKT}}                          & \colorbox{myblue}{0.7058}            & 0.4107            & \colorbox{myred}{0}  & 0.6736           & 0.4674           & \colorbox{myred}{0}            & 0.6990            & 0.3613            & \colorbox{myred}{0}            & 0.6633            & 0.4129           & \colorbox{myred}{0}    & 0.8870           & 0.3558           & \colorbox{myred}{0}      \\
\multicolumn{1}{l|}{+FFT}                         & \colorbox{mygreen}{0.7063}      & \colorbox{mygreen}{0.4071}      & $\geq$17,200 & \colorbox{mygreen}{0.6852}           & \colorbox{mygreen}{0.4633}           & $\geq$9,900 &  \colorbox{mygreen}{0.7066}      & \colorbox{mygreen}{0.3594}      & $\geq$18,300  & \colorbox{mygreen}{0.7116}      & \colorbox{mygreen}{0.3992}           & $\geq$33,600       & \colorbox{mygreen}{0.8873}           & \colorbox{mygreen}{0.3555}           & \colorbox{myblue}{$\geq$4,200}     \\
\multicolumn{1}{l|}{+Adapter}                     & 0.6749            & 0.4242            & $\geq$16,600 & \colorbox{myblue}{0.6783}           & 0.4661           & $\geq$9,800 & 0.6634            & 0.3714            & $\geq$17,700  & 0.6467            & 0.4275           & $\geq$36,100       & 0.8851           & 0.3575           & $\geq$4,500     \\
\multicolumn{1}{l|}{+BitFit}                      & 0.7054            & \colorbox{myblue}{0.4080}            & \colorbox{myblue}{$\geq$16,300} & 0.6770           & \colorbox{myblue}{0.4655}           & \colorbox{myblue}{$\geq$9,500} & \colorbox{myblue}{0.7039}            & \colorbox{myblue}{0.3599}            & \colorbox{myblue}{$\geq$14,100}  & \colorbox{myblue}{0.6841}            & \colorbox{myblue}{0.4105}           & \colorbox{myblue}{$\geq$32,600}     & \colorbox{myblue}{0.8871}           & \colorbox{myblue}{0.3556}           & $\geq$4,400       \\
\multicolumn{1}{l|}{\textbf{+Cuff-KT}}               & \cellcolor{gray!40}\colorbox{myred}{\textbf{0.8130**}} & \cellcolor{gray!40}\colorbox{myred}{\textbf{0.3773**}}  & \cellcolor{gray!40}\colorbox{mygreen}{\textbf{$\geq$419}}  & \cellcolor{gray!40}\colorbox{myred}{\textbf{0.7176**}}           & \cellcolor{gray!40}\colorbox{myred}{\textbf{0.4573**}}           & \cellcolor{gray!40}\colorbox{mygreen}{\textbf{$\geq$164}}   & \cellcolor{gray!40}\colorbox{myred}{\textbf{0.7834**}} & \cellcolor{gray!40}\colorbox{myred}{\textbf{0.3459**}} & \cellcolor{gray!40}\colorbox{mygreen}{\textbf{$\geq$435}}     & \cellcolor{gray!40}\colorbox{myred}{\textbf{0.7176**}}   & \cellcolor{gray!40}\colorbox{myred}{\textbf{0.3931**}}  & \cellcolor{gray!40}\colorbox{mygreen}{\textbf{$\geq$1,211}}     & \cellcolor{gray!40}\colorbox{myred}{\textbf{0.8979**}}           & \cellcolor{gray!40}\colorbox{myred}{\textbf{0.3493**}}           & \multicolumn{1}{c}{\cellcolor{gray!40}\colorbox{mygreen}{\textbf{$\geq$76}}}       \\ \midrule

\rowcolor[HTML]{F8F8F8}\multicolumn{1}{l|}{\cellcolor{white}\textcolor[rgb]{1,0.8,0}{DKVMN} $^{\dagger}$}                          & 0.6727            & 0.4194            & \colorbox{myred}{0}  & 0.6464           & 0.4765           & \colorbox{myred}{0}            & 0.6851            & 0.3640            & \colorbox{myred}{0}            & 0.6027            & 0.4313           & \colorbox{myred}{0}    & 0.7730           & 0.4225           & \colorbox{myred}{0}      \\
\multicolumn{1}{l|}{+FFT}                         & \colorbox{myblue}{0.6748}      & \colorbox{mygreen}{0.4144}      & \colorbox{myblue}{$\geq$47,000} & \colorbox{mygreen}{0.6597}           & \colorbox{mygreen}{0.4714}           & $\geq$35,700 &  \colorbox{mygreen}{0.6956}      & \colorbox{mygreen}{0.3623}      & $\geq$52,000  & \colorbox{mygreen}{0.6530}      & \colorbox{mygreen}{0.4124}           & $\geq$92,800      & \colorbox{mygreen}{0.7756}        & \colorbox{mygreen}{0.4211}           & $\geq$17,800     \\
\multicolumn{1}{l|}{+Adapter}                     & 0.6728            & 0.4148           & $\geq$63,600 & 0.6508          & \colorbox{myblue}{0.4742}           & $\geq$34,800 & \colorbox{myblue}{0.6908}            & 0.3633          & $\geq$48,800  & \colorbox{myblue}{0.6283}            & \colorbox{myblue}{0.4161}          & \colorbox{myblue}{$\geq$78,800}       & 0.7702           & 0.4237           & $\geq$16,900     \\
\multicolumn{1}{l|}{+BitFit}                      & \colorbox{mygreen}{0.6756}           & \colorbox{myblue}{0.4146}            & $\geq$60,500 & \colorbox{myblue}{0.6524}          & 0.4743        & \colorbox{myblue}{$\geq$32,000} & 0.6900            & \colorbox{myblue}{0.3631}            & \colorbox{myblue}{$\geq$45,200}  & 0.6213            & 0.4184          & $\geq$80,500     & \colorbox{myblue}{0.7742}           & \colorbox{myblue}{0.4217}           & \colorbox{myblue}{$\geq$15,800}       \\
\multicolumn{1}{l|}{\textbf{+Cuff-KT}}               & \cellcolor{gray!40}\colorbox{myred}{\textbf{0.8433**}} & \cellcolor{gray!40}\colorbox{myred}{\textbf{0.3664**}}  & \cellcolor{gray!40}\colorbox{mygreen}{\textbf{$\geq$171}}  & \cellcolor{gray!40}\colorbox{myred}{\textbf{0.7325**}}   & \cellcolor{gray!40}\colorbox{myred}{\textbf{0.4612**}}  & \cellcolor{gray!40}\colorbox{mygreen}{\textbf{$\geq$163}}   & \cellcolor{gray!40}\colorbox{myred}{\textbf{0.7943**}} & \cellcolor{gray!40}\colorbox{myred}{\textbf{0.3425**}} & \cellcolor{gray!40}\colorbox{mygreen}{\textbf{$\geq$362}}     & \cellcolor{gray!40}\colorbox{myred}{\textbf{0.7359**}}   & \cellcolor{gray!40}\colorbox{myred}{\textbf{0.4119}}  & \cellcolor{gray!40}\colorbox{mygreen}{\textbf{$\geq$536}}     & \cellcolor{gray!40}\colorbox{myred}{\textbf{0.8901**}}   & \cellcolor{gray!40}\colorbox{myred}{\textbf{0.3563**}}  & \multicolumn{1}{c}{\cellcolor{gray!40}\colorbox{mygreen}{\textbf{$\geq$136}}}       \\ \midrule

\rowcolor[HTML]{F8F8F8}\multicolumn{1}{l|}{\cellcolor{white}\textcolor{blue}{AT-DKT}}                        & 0.6981            & 0.4106            & \colorbox{myred}{0}    & 0.6924           & 0.4625           & \colorbox{myred}{0}        & 0.6922            & 0.3621            & \colorbox{myred}{0}            & 0.6437            & 0.4228           & \colorbox{myred}{0}       & 0.8854           & 0.3572           & \colorbox{myred}{0}   \\
\multicolumn{1}{l|}{+FFT}                         & \colorbox{mygreen}{0.7005}      & \colorbox{mygreen}{0.4083}      & $\geq$126,400 & \colorbox{mygreen}{0.7061}           & \colorbox{mygreen}{0.4576}           & $\geq$62,200 & \colorbox{mygreen}{0.7020}      & \colorbox{mygreen}{0.3602}      & $\geq$95,000  & \colorbox{mygreen}{0.6918}      & \colorbox{mygreen}{0.4068}     & $\geq$176,000     & \colorbox{mygreen}{0.8874}           & \colorbox{mygreen}{0.3556}           & $\geq$31,500       \\
\multicolumn{1}{l|}{+Adapter}                     & 0.6588            & 0.4287            & $\geq$125,100 & 0.6938           & 0.4609           & \colorbox{myblue}{$\geq$59,700} & 0.6443            & 0.3878            & \colorbox{myblue}{$\geq$88,800}  & 0.6276            & 0.4351           & \colorbox{myblue}{$\geq$168,300}      & 0.8829           & 0.3591           & \colorbox{myblue}{$\geq$30,800}      \\
\multicolumn{1}{l|}{+BitFit}                      & \colorbox{myblue}{0.6989}            & \colorbox{myblue}{0.4094}            & \colorbox{myblue}{$\geq$121,300} & \colorbox{myblue}{0.6973}           & \colorbox{myblue}{0.4598}           & $\geq$60,000 & \colorbox{myblue}{0.6990}            & \colorbox{myblue}{0.3608}            & $\geq$91,300  & \colorbox{myblue}{0.6668}            & \colorbox{myblue}{0.4178}           & $\geq$169,300     & \colorbox{myblue}{0.8866}           & \colorbox{myblue}{0.3562}           & $\geq$31,300       \\
\multicolumn{1}{l|}{\textbf{+Cuff-KT}}               & \cellcolor{gray!40}\colorbox{myred}{\textbf{0.8335**}} & \cellcolor{gray!40}\colorbox{myred}{\textbf{0.3714**}} & \cellcolor{gray!40}\colorbox{mygreen}{\textbf{$\geq$236}}  & \cellcolor{gray!40}\colorbox{myred}{\textbf{0.7322**}}           & \cellcolor{gray!40}\colorbox{myred}{\textbf{0.4560*}}           & \cellcolor{gray!40}\colorbox{mygreen}{\textbf{$\geq$228}}    & \cellcolor{gray!40}\colorbox{myred}{\textbf{0.7869**}} & \cellcolor{gray!40}\colorbox{myred}{\textbf{0.3435**}} & \cellcolor{gray!40}\colorbox{mygreen}{\textbf{$\geq$254}}     & \cellcolor{gray!40}\colorbox{myred}{\textbf{0.7133**}} & \cellcolor{gray!40}\colorbox{myred}{\textbf{0.4009*}} & \cellcolor{gray!40}\colorbox{mygreen}{\textbf{$\geq$784}}      & \cellcolor{gray!40}\colorbox{myred}{\textbf{0.8960**}}           & \cellcolor{gray!40}\colorbox{myred}{\textbf{0.3517**}}           &  \multicolumn{1}{c}{\cellcolor{gray!40}\colorbox{mygreen}{\textbf{$\geq$119}}}     \\ \midrule

\rowcolor[HTML]{F8F8F8}\multicolumn{1}{l|}{\cellcolor{white}\textcolor[rgb]{0.0,0.9,0.5}{stableKT} $^{\dagger}$}                          & 0.6725            & 0.4172            & \colorbox{myred}{0}  & 0.6892           & 0.4650           & \colorbox{myred}{0}            & 0.6909            & 0.3633            & \colorbox{myred}{0}            & 0.6417            & 0.4349           & \colorbox{myred}{0}    & 0.7805           & 0.4204           & \colorbox{myred}{0}      \\
\multicolumn{1}{l|}{+FFT}                         & \colorbox{myblue}{0.6745}      & \colorbox{myblue}{0.4141}      & $\geq$113,900 & \colorbox{mygreen}{0.7031}           & \colorbox{mygreen}{0.4599}           & $\geq$57,500 &  \colorbox{myblue}{0.6945}      & \colorbox{myblue}{0.3621}      & $\geq$84,900  & \colorbox{mygreen}{0.6790}      & \colorbox{myred}{0.4130}           & $\geq$149,200      & \colorbox{myblue}{0.7825}        & \colorbox{myblue}{0.4190}           & \colorbox{myblue}{$\geq$29,600}     \\
\multicolumn{1}{l|}{+Adapter}                     & 0.6742            & 0.4144           & $\geq$130,700 & 0.6943          & 0.4612           & $\geq$66,600 & 0.6917            & 0.3629          & $\geq$97,200  & \colorbox{myblue}{0.6652}            & \colorbox{mygreen}{0.4166}          & $\geq$168,800       & 0.7795           & 0.4201           & $\geq$36,800     \\
\multicolumn{1}{l|}{+BitFit}                      & \colorbox{mygreen}{0.6780}           & \colorbox{mygreen}{0.4135}            & \colorbox{myblue}{$\geq$110,800} & \colorbox{myblue}{0.6968}          & \colorbox{myblue}{0.4601}        & \colorbox{myblue}{$\geq$55,900} & \colorbox{mygreen}{0.6949}            & \colorbox{mygreen}{0.3620}            & \colorbox{myblue}{$\geq$79,000}  & 0.6615            & \colorbox{myblue}{0.4174}          & \colorbox{myblue}{$\geq$143,100}     & \colorbox{mygreen}{0.7846}           & \colorbox{mygreen}{0.4172}           & $\geq$30,700       \\
\multicolumn{1}{l|}{\textbf{+Cuff-KT}}               & \cellcolor{gray!40}\colorbox{myred}{\textbf{0.7864**}} & \cellcolor{gray!40}\colorbox{myred}{\textbf{0.3895**}}  & \cellcolor{gray!40}\colorbox{mygreen}{\textbf{$\geq$583}}  & \cellcolor{gray!40}\colorbox{myred}{\textbf{0.7272**}}   & \cellcolor{gray!40}\colorbox{myred}{\textbf{0.4545*}}  & \cellcolor{gray!40}\colorbox{mygreen}{\textbf{$\geq$367}}   & \cellcolor{gray!40}\colorbox{myred}{\textbf{0.7506**}} & \cellcolor{gray!40}\colorbox{myred}{\textbf{0.3546**}} & \cellcolor{gray!40}\colorbox{mygreen}{\textbf{$\geq$631}}     & \cellcolor{gray!40}\colorbox{myred}{\textbf{0.7149**}}   & \cellcolor{gray!40}\colorbox{mygreen}{\textbf{0.4166}}  & \cellcolor{gray!40}\colorbox{mygreen}{\textbf{$\geq$875}}     & \cellcolor{gray!40}\colorbox{myred}{\textbf{0.8942**}}   & \cellcolor{gray!40}\colorbox{myred}{\textbf{0.3529**}}  & \multicolumn{1}{c}{\cellcolor{gray!40}\colorbox{mygreen}{\textbf{$\geq$208}}}       \\ \midrule
\rowcolor[HTML]{F8F8F8}\multicolumn{1}{l|}{\cellcolor{white}\textcolor[rgb]{0.54,0.17,0.89}{DIMKT}}                        & 0.7055            & 0.4080            & \colorbox{myred}{0}    & 0.7321           & 0.4512           & \colorbox{myred}{0}        & 0.7934            & 0.3404            & \colorbox{myred}{0}            & 0.8322            & 0.3402           & \colorbox{myred}{0}        & 0.8911           & 0.3523           & \colorbox{myred}{0}  \\
\multicolumn{1}{l|}{+FFT}                         & \colorbox{myblue}{0.7072}            & \colorbox{myblue}{0.4063}            & $\geq$270,900 & \colorbox{mygreen}{0.7451}           & \colorbox{myred}{0.4449}           & $\geq$125,400 & \colorbox{mygreen}{0.8000}      & \colorbox{mygreen}{0.3375}      & $\geq$205,200 & \colorbox{myblue}{0.8366}            & \colorbox{myblue}{0.3383}           & $\geq$377,800    & \colorbox{myblue}{0.8916}           & \colorbox{myblue}{0.3515}           & $\geq$63,100        \\
\multicolumn{1}{l|}{+Adapter}                     & 0.6507            & 0.4387            & $\geq$410,000 & 0.7386           & \colorbox{myblue}{0.4465}           & $\geq$171,200 & 0.7526            & 0.3671            & $\geq$278,500 & 0.7929            & 0.3696           & $\geq$509,200       & 0.8893           & 0.3532           & $\geq$85,200     \\
\multicolumn{1}{l|}{+BitFit}                      & \colorbox{mygreen}{0.7082}      & \colorbox{mygreen}{0.4061}      & \colorbox{myblue}{$\geq$263,500} & \colorbox{myblue}{0.7387}           & \colorbox{myblue}{0.4465}           & \colorbox{myblue}{$\geq$119,900} & \colorbox{myblue}{0.7972}            & \colorbox{myblue}{0.3382}            & \colorbox{myblue}{$\geq$199,400} & \colorbox{mygreen}{0.8369}      & \colorbox{mygreen}{0.3381}     & \colorbox{myblue}{$\geq$347,800}       & \colorbox{mygreen}{0.8919}           & \colorbox{mygreen}{0.3511}           & \colorbox{myblue}{$\geq$56,100}     \\
\multicolumn{1}{l|}{\textbf{+Cuff-KT}}                                      & \cellcolor{gray!40}\colorbox{myred}{\textbf{0.8322**}} & \cellcolor{gray!40}\colorbox{myred}{\textbf{0.3710*}}  & \cellcolor{gray!40}\colorbox{mygreen}{\textbf{$\geq$232}}  & \cellcolor{gray!40}\colorbox{myred}{\textbf{0.7509**}}           & \cellcolor{gray!40}\colorbox{mygreen}{\textbf{0.4450}}           & \cellcolor{gray!40}\colorbox{mygreen}{\textbf{$\geq$166}}   & \cellcolor{gray!40}\colorbox{myred}{\textbf{0.8380**}} & \cellcolor{gray!40}\colorbox{myred}{\textbf{0.3297**}} & \cellcolor{gray!40}\colorbox{mygreen}{\textbf{$\geq$275}}     & \cellcolor{gray!40}\colorbox{myred}{\textbf{0.8540*}}  & \cellcolor{gray!40}\colorbox{myred}{\textbf{0.3347*}} & \cellcolor{gray!40}\colorbox{mygreen}{\textbf{$\geq$339}}       & \cellcolor{gray!40}\colorbox{myred}{\textbf{0.9034*}}           & \cellcolor{gray!40}\colorbox{myred}{\textbf{0.3439*}}           & \multicolumn{1}{c}{\cellcolor{gray!40}\colorbox{mygreen}{\textbf{$\geq$83}}}     \\ 
\bottomrule[1.5pt]
\end{tabular}
}
\end{table*}

\begin{table*}[h]
\centering
\caption{Performance comparison between different methods under inter-learner shift.}
\vspace{-0.4cm}
\label{tab: result_inter-learner}
\resizebox{\textwidth}{!}{%
\begin{tabular}{@{}cc|c|c|c|c|c|c|c|c|c|c|c|c|c|c@{}}
\toprule[1.5pt]
\multicolumn{1}{c|}{\textbf{Dataset}}                                           & \multicolumn{3}{c|}{\textbf{assist15}} & \multicolumn{3}{c|}{\textbf{assist17}} & \multicolumn{3}{c|}{\textbf{comp}}     & \multicolumn{3}{c|}{\textbf{xes3g5m}} & \multicolumn{3}{c}{\textbf{dbe-kt22}}  \\ \cmidrule(l){1-1} \cmidrule(l){2-4} \cmidrule(l){5-7} \cmidrule(l){8-10} \cmidrule(l){11-13} \cmidrule(l){14-16}
\multicolumn{1}{c|}{\textbf{Metric}} & \textbf{AUC $\uparrow$}    & \textbf{RMSE $\downarrow$}   & \textbf{TO (ms) $\downarrow$} & \textbf{AUC $\uparrow$}    & \textbf{RMSE $\downarrow$}   & \textbf{TO (ms) $\downarrow$} & \textbf{AUC $\uparrow$}    & \textbf{RMSE $\downarrow$}   & \textbf{TO (ms) $\downarrow$} & \textbf{AUC $\uparrow$}    & \textbf{RMSE $\downarrow$}   & \textbf{TO (ms) $\downarrow$}  & \textbf{AUC $\uparrow$}    & \textbf{RMSE $\downarrow$}   & \textbf{TO (ms) $\downarrow$} \\ \cmidrule(l){1-16} 
\rowcolor[HTML]{F8F8F8}\multicolumn{1}{l|}{\cellcolor{white}\textcolor[rgb]{1,0,0}{DKT}}                          & 0.7075            & 0.4363           & \colorbox{myred}{0}      & 0.6585            & 0.4711           & \colorbox{myred}{0}     & 0.6681            & 0.4355           & \colorbox{myred}{0}          & 0.7907           & 0.4329           & \colorbox{myred}{0}      & \colorbox{myblue}{0.9274}            & 0.3014           & \colorbox{myred}{0}    \\
\multicolumn{1}{l|}{+FFT}                         & \colorbox{mygreen}{0.7137}      & \colorbox{mygreen}{0.4339}     & $\geq$18,800   & \colorbox{mygreen}{0.6636}            & \colorbox{mygreen}{0.4670}           &  $\geq$2,100  & \colorbox{mygreen}{0.6839}      & \colorbox{mygreen}{0.4310}     & $\geq$3,600      & \colorbox{mygreen}{0.7990}     & \colorbox{mygreen}{0.4166}     & $\geq$4,400      & \colorbox{mygreen}{0.9278}            & \colorbox{mygreen}{0.3005}           & \colorbox{myblue}{$\geq$1,300}      \\
\multicolumn{1}{l|}{+Adapter}                     & 0.6805            & 0.4456           & \colorbox{myblue}{$\geq$17,000}   & 0.6445            & 0.4768           & $\geq$24,900   & 0.6461            & 0.4438           & $\geq$3,200      & 0.7646           & 0.4427           & \colorbox{myblue}{$\geq$4,300}     & 0.9254            & 0.3033           & $\geq$19,800       \\
\multicolumn{1}{l|}{+BitFit}                     & \colorbox{myblue}{0.7119}            & \colorbox{myblue}{0.4349}           & $\geq$17,200   & \colorbox{myblue}{0.6590}            & \colorbox{myblue}{0.4703}           & \colorbox{myblue}{$\geq$1,400}   & \colorbox{myblue}{0.6734}            & \colorbox{myblue}{0.4326}           & \colorbox{myblue}{$\geq$3,100}      & \colorbox{myblue}{0.7905}           & \colorbox{myblue}{0.4323}          & $\geq$4,900   & \colorbox{myblue}{0.9274}            & \colorbox{myblue}{0.3010}           & $\geq$1,400         \\
\multicolumn{1}{l|}{\textbf{+Cuff-KT}}              & \cellcolor{gray!40}\colorbox{myred}{\textbf{0.7365*}}  & \cellcolor{gray!40}\colorbox{myred}{\textbf{0.4302}}  & \cellcolor{gray!40}\colorbox{mygreen}{\textbf{$\geq$355}}     & \cellcolor{gray!40}\colorbox{myred}{\textbf{0.6851**}} & \cellcolor{gray!40}\colorbox{myred}{\textbf{0.4654*}}  & \cellcolor{gray!40}\colorbox{mygreen}{\textbf{$\geq$73}}    & \cellcolor{gray!40}\colorbox{myred}{\textbf{0.6937**}} & \cellcolor{gray!40}\colorbox{myred}{\textbf{0.4294*}} & \cellcolor{gray!40}\colorbox{mygreen}{\textbf{$\geq$96}}         & \cellcolor{gray!40}\colorbox{myred}{\textbf{0.8004}}  & \cellcolor{gray!40}\colorbox{myred}{\textbf{0.4158}}  & \cellcolor{gray!40}\colorbox{mygreen}{\textbf{$\geq$123}}    & \cellcolor{gray!40}\colorbox{myred}{\textbf{0.9342*}} & \cellcolor{gray!40}\colorbox{myred}{\textbf{0.2964}}  & \multicolumn{1}{c}{\cellcolor{gray!40}\colorbox{mygreen}{\textbf{$\geq$62}}}         \\ \midrule

\rowcolor[HTML]{F8F8F8}\multicolumn{1}{l|}{\cellcolor{white}\textcolor[rgb]{1,0.8,0}{DKVMN} $^{\dagger}$}                          & 0.6885            & 0.4413            & \colorbox{myred}{0}  & 0.6484           & 0.4765           & \colorbox{myred}{0}            & 0.6567            & 0.4426            & \colorbox{myred}{0}            & \colorbox{myblue}{0.7207}            & 0.4595           & \colorbox{myred}{0}    & \colorbox{mygreen}{0.8082}           & \colorbox{mygreen}{0.3836}           & \colorbox{myred}{0}      \\
\multicolumn{1}{l|}{+FFT}                         & \colorbox{mygreen}{0.6974}      & \colorbox{mygreen}{0.4389}      & $\geq$45,100 & \colorbox{mygreen}{0.6505}           & \colorbox{mygreen}{0.4738}           & $\geq$8,800 &  \colorbox{mygreen}{0.6661}      & \colorbox{mygreen}{0.4347}      & $\geq$15,600  & \colorbox{mygreen}{0.7316}      & \colorbox{mygreen}{0.4451}           & $\geq$17,000      & 0.8058        & \colorbox{myblue}{0.3837}           & $\geq$7,700     \\
\multicolumn{1}{l|}{+Adapter}                     & 0.6937            & \colorbox{myblue}{0.4399}           & \colorbox{myblue}{$\geq$40,700} & 0.6299          & 0.4838           & $\geq$46,500 & 0.6517            & \colorbox{myblue}{0.4403}          & \colorbox{myblue}{$\geq$14,300}  & 0.7181            & \colorbox{myblue}{0.4540}          & $\geq$25,300       & 0.8004           & 0.3864           & $\geq$19,200     \\
\multicolumn{1}{l|}{+BitFit}                      & \colorbox{myblue}{0.6938}           & 0.4402            & $\geq$45,600 & \colorbox{myblue}{0.6494}          & \colorbox{myblue}{0.4751}        & \colorbox{myblue}{$\geq$7,600} & \colorbox{myblue}{0.6572}            & 0.4416            & $\geq$20,200  & 0.7205            & 0.4590          & \colorbox{myblue}{$\geq$13,000}     & \colorbox{myblue}{0.8076}           & \colorbox{mygreen}{0.3836}           & \colorbox{myblue}{$\geq$6,700}       \\
\multicolumn{1}{l|}{\textbf{+Cuff-KT}}               & \cellcolor{gray!40}\colorbox{myred}{\textbf{0.7284**}} & \cellcolor{gray!40}\colorbox{myred}{\textbf{0.4340*}}  & \cellcolor{gray!40}\colorbox{mygreen}{\textbf{$\geq$428}}  & \cellcolor{gray!40}\colorbox{myred}{\textbf{0.6805**}}   & \cellcolor{gray!40}\colorbox{myred}{\textbf{0.4685*}}  & \cellcolor{gray!40}\colorbox{mygreen}{\textbf{$\geq$91}}   & \cellcolor{gray!40}\colorbox{myred}{\textbf{0.6917**}} & \cellcolor{gray!40}\colorbox{myred}{\textbf{0.4334}} & \cellcolor{gray!40}\colorbox{mygreen}{\textbf{$\geq$121}}     & \cellcolor{gray!40}\colorbox{myred}{\textbf{0.7896**}}   & \cellcolor{gray!40}\colorbox{myred}{\textbf{0.4214**}}  & \cellcolor{gray!40}\colorbox{mygreen}{\textbf{$\geq$197}}     & \cellcolor{gray!40}\colorbox{myred}{\textbf{0.9256**}}   & \cellcolor{gray!40}\colorbox{myred}{\textbf{0.3042**}}  & \multicolumn{1}{c}{\cellcolor{gray!40}\colorbox{mygreen}{\textbf{$\geq$83}}}       \\ \midrule

\rowcolor[HTML]{F8F8F8}\multicolumn{1}{l|}{\cellcolor{white}\textcolor{blue}{AT-DKT}}                        & 0.7030            & 0.4389           & \colorbox{myred}{0}      & 0.6596            & 0.4692           & \colorbox{myred}{0}     & 0.6587            & 0.4375           & \colorbox{myred}{0}          & \colorbox{myblue}{0.7868}           & 0.4370           & \colorbox{myred}{0}     & \colorbox{myblue}{0.9228}            & \colorbox{myblue}{0.3069}           & \colorbox{myred}{0}     \\
\multicolumn{1}{l|}{+FFT}                        & \colorbox{mygreen}{0.7104}      & \colorbox{mygreen}{0.4355}     & $\geq$74,700   & \colorbox{mygreen}{0.6631}            & \colorbox{mygreen}{0.4670}           & $\geq$7,800   & \colorbox{mygreen}{0.6751}      & \colorbox{mygreen}{0.4312}     & $\geq$20,300     & \colorbox{mygreen}{0.7916}     & \colorbox{mygreen}{0.4242}     & $\geq$21,300      & 0.9225            & \colorbox{myblue}{0.3069}           & \colorbox{myblue}{$\geq$5,200}     \\
\multicolumn{1}{l|}{+Adapter}                    & 0.6708            & 0.4520           & \colorbox{myblue}{$\geq$55,300}    & 0.6406            & 0.4743           & $\geq$14,500  & 0.6253            & 0.4498           & \colorbox{myblue}{$\geq$18,100}     & 0.7643           & 0.4457           & $\geq$23,500     & 0.9210            & 0.3085           & $\geq$6,800      \\
\multicolumn{1}{l|}{+BitFit}                      & \colorbox{myblue}{0.7076}            & \colorbox{myblue}{0.4367}           & $\geq$59,100    & \colorbox{myblue}{0.6603}            & \colorbox{myblue}{0.4680}           & \colorbox{myblue}{$\geq$6,800}  & \colorbox{myblue}{0.6666}            & \colorbox{myblue}{0.4334}           & $\geq$18,600     & 0.7860           & \colorbox{myblue}{0.4352}           & \colorbox{myblue}{$\geq$19,300}       & \colorbox{mygreen}{0.9231}            & \colorbox{mygreen}{0.3067}           & $\geq$5,900    \\
\multicolumn{1}{l|}{\textbf{+Cuff-KT}}               & \cellcolor{gray!40}\colorbox{myred}{\textbf{0.7348**}} & \cellcolor{gray!40}\colorbox{myred}{\textbf{0.4316*}} & \cellcolor{gray!40}\colorbox{mygreen}{\textbf{$\geq$170}}      & \cellcolor{gray!40}\colorbox{myred}{\textbf{0.6852**}} & \cellcolor{gray!40}\colorbox{myred}{\textbf{0.4652}}  & \cellcolor{gray!40}\colorbox{mygreen}{\textbf{$\geq$46}}   & \cellcolor{gray!40}\colorbox{myred}{\textbf{0.6919**}} & \cellcolor{gray!40}\colorbox{myred}{\textbf{0.4303*}} & \cellcolor{gray!40}\colorbox{mygreen}{\textbf{$\geq$64}}         & \cellcolor{gray!40}\colorbox{myred}{\textbf{0.7959*}} & \cellcolor{gray!40}\colorbox{myred}{\textbf{0.4183*}} & \cellcolor{gray!40}\colorbox{mygreen}{\textbf{$\geq$110}}   & \cellcolor{gray!40}\colorbox{myred}{\textbf{0.9343*}} & \cellcolor{gray!40}\colorbox{myred}{\textbf{0.2960}}  & \multicolumn{1}{c}{\cellcolor{gray!40}\colorbox{mygreen}{\textbf{$\geq$37}}}         \\ \midrule
\rowcolor[HTML]{F8F8F8}\multicolumn{1}{l|}{\cellcolor{white}\textcolor[rgb]{0.0,0.9,0.5}{stableKT} $^{\dagger}$}                          & 0.6928            & 0.4445            & \colorbox{myred}{0}  & \colorbox{mygreen}{0.6785}           & 0.4738           & \colorbox{myred}{0}            & \colorbox{myblue}{0.6618}            & 0.4381            & \colorbox{myred}{0}            & \colorbox{myblue}{0.7424}            & 0.4512           & \colorbox{myred}{0}    & \colorbox{mygreen}{0.8105}           & \colorbox{myblue}{0.3825}           & \colorbox{myred}{0}      \\
\multicolumn{1}{l|}{+FFT}                         & \colorbox{mygreen}{0.6995}      & \colorbox{mygreen}{0.4385}      & $\geq$68,700 & 0.6754           & \colorbox{myred}{0.4663}           & $\geq$11,100 &  \colorbox{mygreen}{0.6731}      & \colorbox{mygreen}{0.4328}      & $\geq$23,100  & \colorbox{mygreen}{0.7494}      & \colorbox{mygreen}{0.4381}           & $\geq$22,800      & 0.8066        & 0.3843           & $\geq$8,300     \\
\multicolumn{1}{l|}{+Adapter}                     & 0.6942            & 0.4398           & $\geq$77,700 & 0.6499          & 0.4794           & $\geq$12,000 & 0.6379            & 0.4479          & $\geq$23,400  & 0.7277            & 0.4624          & $\geq$23,400       & 0.8026           & 0.3853           & $\geq$14,100     \\
\multicolumn{1}{l|}{+BitFit}                      & \colorbox{myblue}{0.6969}           & \colorbox{myblue}{0.4393}            & \colorbox{myblue}{$\geq$65,700} & \colorbox{myblue}{0.6783}          & \colorbox{myblue}{0.4732}        & \colorbox{myblue}{$\geq$8,900} & 0.6616            & \colorbox{myblue}{0.4374}            & \colorbox{myblue}{$\geq$19,500}  & 0.7419            & \colorbox{myblue}{0.4498}          & \colorbox{myblue}{$\geq$19,400}     & \colorbox{myblue}{0.8098}           & \colorbox{mygreen}{0.3822}           & \colorbox{myblue}{$\geq$11,100}       \\
\multicolumn{1}{l|}{\textbf{+Cuff-KT}}               & \cellcolor{gray!40}\colorbox{myred}{\textbf{0.7242**}} & \cellcolor{gray!40}\colorbox{myred}{\textbf{0.4343*}}  & \cellcolor{gray!40}\colorbox{mygreen}{\textbf{$\geq$613}}  & \cellcolor{gray!40}\colorbox{myred}{\textbf{0.6912*}}   & \cellcolor{gray!40}\colorbox{mygreen}{\textbf{0.4681}}  & \cellcolor{gray!40}\colorbox{mygreen}{\textbf{$\geq$118}}   & \cellcolor{gray!40}\colorbox{myred}{\textbf{0.6877*}} & \cellcolor{gray!40}\colorbox{myred}{\textbf{0.4301}} & \cellcolor{gray!40}\colorbox{mygreen}{\textbf{$\geq$156}}     & \cellcolor{gray!40}\colorbox{myred}{\textbf{0.7904**}}   & \cellcolor{gray!40}\colorbox{myred}{\textbf{0.4227**}}  & \cellcolor{gray!40}\colorbox{mygreen}{\textbf{$\geq$223}}     & \cellcolor{gray!40}\colorbox{myred}{\textbf{0.9271**}}   & \cellcolor{gray!40}\colorbox{myred}{\textbf{0.3046**}}  & \multicolumn{1}{c}{\cellcolor{gray!40}\colorbox{mygreen}{\textbf{$\geq$107}}}
\\ \midrule
\rowcolor[HTML]{F8F8F8}\multicolumn{1}{l|}{\cellcolor{white}\textcolor[rgb]{0.54,0.17,0.89}{DIMKT}}                        & 0.7134            & 0.4350           & \colorbox{myred}{0}     & \colorbox{myblue}{0.7282}            & \colorbox{myblue}{0.4530}           & \colorbox{myred}{0}      & 0.7556            & 0.4118           & \colorbox{myred}{0}          & \colorbox{myblue}{0.8255}           & 0.4088           & \colorbox{myred}{0}      & \colorbox{mygreen}{0.9339}            & 0.2958           & \colorbox{myred}{0}    \\
\multicolumn{1}{l|}{+FFT}                         & \colorbox{mygreen}{0.7187}      & \colorbox{mygreen}{0.4320}     & $\geq$173,500  & \colorbox{myred}{0.7305}            & \colorbox{myred}{0.4492}           & $\geq$23,500   & \colorbox{mygreen}{0.7590}      & \colorbox{mygreen}{0.4097}     & $\geq$52,500     & \colorbox{myred}{0.8329}  & \colorbox{myred}{0.3983}  & \colorbox{myblue}{$\geq$48,200}       & \colorbox{myblue}{0.9333}            & \colorbox{myblue}{0.2956}           & $\geq$20,600   \\
\multicolumn{1}{l|}{+Adapter}                   & 0.6648            & 0.4577           & $\geq$217,100   & 0.7054            & 0.4593           & $\geq$35,900  & 0.7017            & 0.4465           & $\geq$82,500     & 0.7467           & 0.4618           & $\geq$81,800       & 0.9313            & 0.2970           & $\geq$34,400    \\
\multicolumn{1}{l|}{+BitFit}                      & \colorbox{myblue}{0.7144}            & \colorbox{myblue}{0.4334}           & \colorbox{myblue}{$\geq$154,400}   & \colorbox{mygreen}{0.7284}            & \colorbox{mygreen}{0.4520}           & \colorbox{myblue}{$\geq$22,000}  & \colorbox{myblue}{0.7563}            & \colorbox{myblue}{0.4110}           & \colorbox{myblue}{$\geq$50,800}     & 0.8254           & \colorbox{myblue}{0.4084}           & $\geq$48,600       & \colorbox{mygreen}{0.9339}            & \colorbox{mygreen}{0.2952}           & \colorbox{myblue}{$\geq$19,300}   \\
\multicolumn{1}{l|}{\textbf{+Cuff-KT}}                                     & \cellcolor{gray!40}\colorbox{myred}{\textbf{0.7425**}} & \cellcolor{gray!40}\colorbox{myred}{\textbf{0.4296}}  & \cellcolor{gray!40}\colorbox{mygreen}{\textbf{$\geq$203}}    & \cellcolor{gray!40}{\textbf{0.7275}} & \cellcolor{gray!40}{\textbf{0.4541}}  & \cellcolor{gray!40}\colorbox{mygreen}{\textbf{$\geq$53}}       & \cellcolor{gray!40}\colorbox{myred}{\textbf{0.7657**}} & \cellcolor{gray!40}\colorbox{myred}{\textbf{0.4057*}} & \cellcolor{gray!40}\colorbox{mygreen}{\textbf{$\geq$64}}      & \cellcolor{gray!40}\colorbox{mygreen}{\textbf{0.8309}}     & \cellcolor{gray!40}\colorbox{mygreen}{\textbf{0.4009}}     & \cellcolor{gray!40}\colorbox{mygreen}{\textbf{$\geq$72}}     & \cellcolor{gray!40}\colorbox{myred}{\textbf{0.9396*}} & \cellcolor{gray!40}\colorbox{myred}{\textbf{0.2902*}}  & \multicolumn{1}{c}{\cellcolor{gray!40}\colorbox{mygreen}{\textbf{$\geq$41}}}   \\  \bottomrule[1.5pt]
\end{tabular}
}
\vspace{-0.2cm}
\end{table*}

\begin{figure*}[t]
    \centering
    \includegraphics[width=\linewidth]{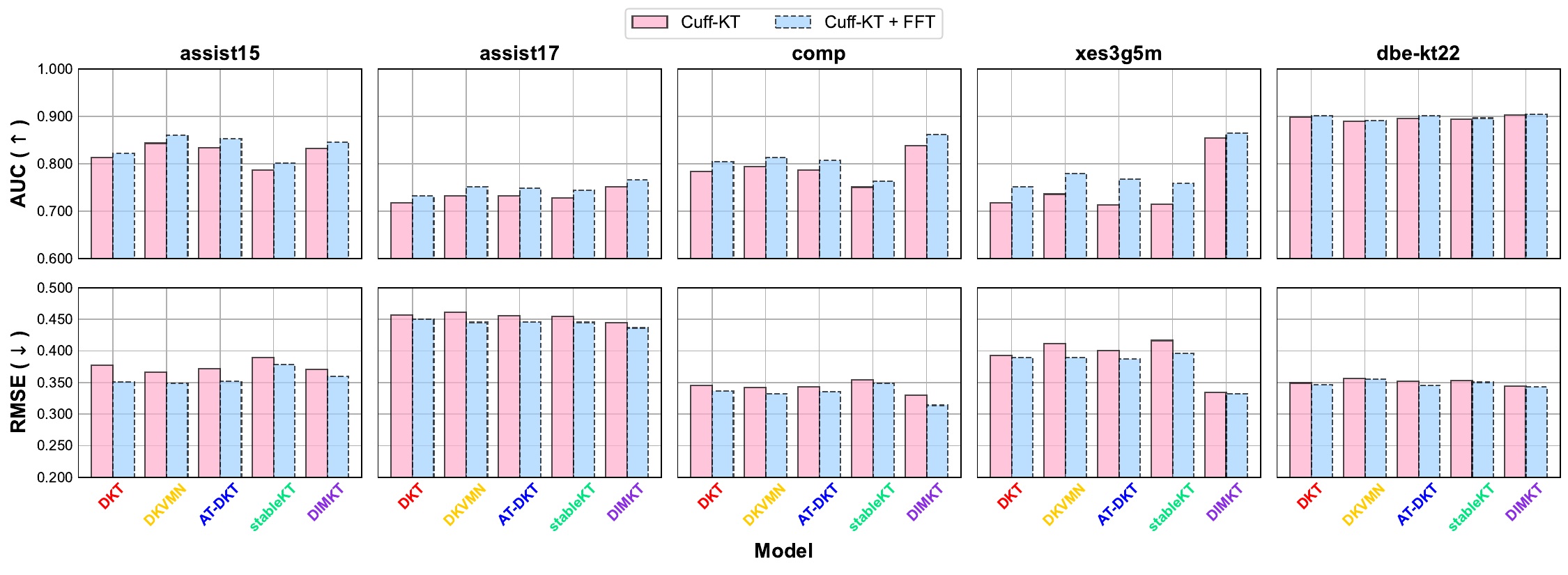}
    \vspace{-0.9cm}
    \caption{The performance of Cuff-KT + FFT under intra-learner shift. The results under inter-learner shift are in Appendix~\ref{sec: cuff+_inter}.}
    \vspace{-0.3cm}
    \label{fig: cuff+}
\end{figure*}

\begin{figure*}[t]
    \centering
    \includegraphics[width=\linewidth]{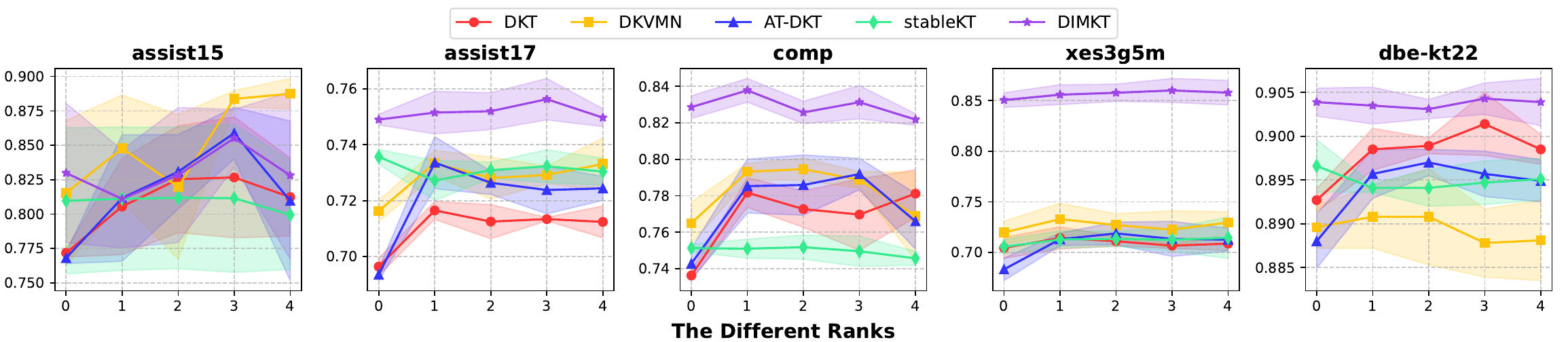}
    \vspace{-0.7cm}
    \caption{Performance in AUC for different ranks under intra-learner shift. The corresponding parameter sizes of the generator in Cuff-KT are shown in Appendix~\ref{sec: parameter_size}.}
    \vspace{-0.3cm}
    \label{fig: rank}
\end{figure*}

\begin{sloppypar}
Under this setting, the generator in Cuff-KT generates parameters for all learners independently of the controller. Tables~\ref{tab: result_intra-learner} and~\ref{tab: result_inter-learner} show the performance comparison between different methods under intra-learner shift and inter-learner shift. Each row is colored as the \colorbox{myred}{\textbf{best}}, \colorbox{mygreen}{\textbf{second best}}, and \colorbox{myblue}{\textbf{third best}}. \textcolor[rgb]{1,0,0}{Different} \textcolor[rgb]{1,0.8,0}{colors} \textcolor{blue}{represent} \textcolor[rgb]{0.0,0.9,0.5}{different} \textcolor[rgb]{0.54,0.17,0.89}{structures}, as mentioned in Sec.~\ref{sec:related_work}. Overall, our Cuff-KT effectively tackles the RLPA task with significant advantages. We can observe:
\begin{itemize}[leftmargin=*, itemsep=0pt, topsep=0pt]
    \item{Overall, compared to baseline methods, our Cuff-KT generally performs best on all metrics over all KT models with different structures across all datasets from various subjects. This performance improvement can be attributed to Cuff-KT's parameter generation approach, which dynamically updates the model to capture distribution dynamics rather than statically considering interaction patterns in the test data, enhancing the KT model's dynamic adaptability.}
    \item{Compared to the backbone, the time overhead caused by Cuff-KT is significantly smaller than fine-tuning-based methods. This is because Cuff-KT updates model parameters only through feedforward computation, without the cumbersome gradient calculations of backpropagation required for retraining.}
    \item{Adapter fine-tuning performs poorly and even leads to performance degradation, as it is heavily affected by task complexity and model scale~\cite{he2021towards, karimi2021compacter}, ultimately resulting in overfitting.}
    \item{Although FFT and BitFit fine-tuning methods generally improve the performance of the backbone, especially FFT based on DKT showing a 0.483 increase in AUC metric on the xes3g5m dataset under intra-learner shift, the time overhead caused is non-negligible in real-world scenarios.}
\end{itemize}
\end{sloppypar}

\subsection{\colorbox{mygray}{\underline{F}lexible} Application}
\label{sec: cuff-kt+}

Thanks to the independence of the generator in our Cuff-KT from fine-tuning-based methods, we attempt to combine Cuff-KT with FFT. The results in terms of AUC and RMSE under intra-learner shifit and inter-learner shift are shown in Figure~\ref{fig: cuff+} and Figure~\ref{apx: cuff+} in Appendix~\ref{sec: cuff+_inter}, respectively. As can be seen from the figures, on different backbone models and across all datasets, the performance still shows a significant improvement after combining Cuff-KT with FFT. This is because FFT can learn different distributions from the recent data, facilitating Cuff-KT's smooth transition to the distribution in the test data. This combination provides a reference for flexibly fine-tuning models in special real-world scenarios where real-time requirements are not high.

Moreover, the generator in Cuff-KT can flexibly generate parameters for or insert into any layer of the KT model. This inspires us to consider how the generator can choose the position and network structure for generation or insertion. Due to space limitations, we leave this as a direction for future research.

\subsection{Ablation Study}
\label{sec: ablation}

\begin{table}
\caption{The performance of different variants in Cuff-KT.}
\vspace{-0.4cm}
\label{tab: ablation}
\centering
\resizebox{0.48\textwidth}{!}{
\begin{tabular}{@{}c|cc|cc|cc|cc|cc@{}}
\toprule[1.5pt]
\textbf{Dataset}       & \multicolumn{2}{c|}{\textbf{assist15}} & \multicolumn{2}{c|}{\textbf{assist17}} & \multicolumn{2}{c|}{\textbf{comp}} & \multicolumn{2}{c|}{\textbf{xes3g5m}} & \multicolumn{2}{c}{\textbf{dbe-kt22}} \\ \cmidrule(l){1-1} \cmidrule(l){2-3} \cmidrule(l){4-5} \cmidrule(l){6-7} \cmidrule(l){8-9} \cmidrule(l){10-11}
Metric        & AUC$\uparrow$          & RMSE$\downarrow$       & AUC$\uparrow$          & RMSE$\downarrow$          & AUC$\uparrow$          & RMSE$\downarrow$        & AUC$\uparrow$          & RMSE$\downarrow$     & AUC$\uparrow$          & RMSE$\downarrow$    \\ \midrule
\rowcolor{skyblue!70}\multicolumn{1}{c}{\textbf{Cuff-KT}}       &  \textbf{0.8130}            &  \textbf{0.3773}    & \textbf{0.7176} & \textbf{0.4573}         &  \textbf{0.7834}          &  \textbf{0.3459}           &  \textbf{0.7176}            &  \textbf{0.3931}        & \textbf{0.8979} & \multicolumn{1}{c}{\textbf{0.3493}}     \\ \midrule
\textbf{w/o.} Dual      &  0.7013            &  0.4126     & 0.6894 & 0.4636        &  0.7245          &  0.3693           &  0.7088            &  0.4094       & 0.8843 & 0.3594     \\
\textbf{w/o.} SFE       &  0.7706            &  0.3925     & 0.7015 & 0.4618        &  0.7204          & 0.3612            &  0.6896            &  0.4140      & 0.8955 & 0.3529      \\
\textbf{w/o.} SAA       &  0.7000            &  0.4141     & 0.6881 & 0.4641        &  0.6877          & 0.3640            &  0.6716            &  0.4212      & 0.8806 & 0.3634      \\
\textbf{w.} SHA       &  0.7810            &  0.3844      & 0.6901 & 0.4639       &  0.6924          & 0.3629            &  0.6767            &  0.4185      & 0.8822 & 0.3602      \\
 \bottomrule[1.5pt]
\end{tabular}
}
\vspace{-0.35cm}
\end{table}

We systematically examine the impact of key components in Cuff-KT based on DKT by constructing four variants under intra-learner shift. ``\textbf{w/o.} Dual" indicates that question and response embeddings are fused (\textit{e.g.}, by summation) after embedding. ``\textbf{w/o.} SFE" means the SFE component is omitted, ``\textbf{w/o.} SAA" means omitting the SAA component, and ``\textbf{w.} SHA" means SAA is replaced with standard multi-head attention. From Table~\ref{tab: ablation}, we can easily observe: (1) Cuff-KT outperforms all variants, especially when the SAA component is removed, the predictive performance generally decreases the most, while Cuff-KT with standard multi-head attention comes next, empirically validating that our designed SAA component can effectively achieve adaptive generalization. (2) Cuff-KT's performance is very low when the SFE component is removed or dual modeling is not employed. We believe this is because Cuff-KT can successfully extract question features and learner response features and effectively learn the difficulty distribution of current questions and the ability distribution of learners based on real-time data.

Additionally, we study the effects of different ranks under intra-learner shift. The performance in AUC of different ranks under intra-learner shift and the parameter size of the generator in Cuff-KT are shown in Figure~\ref{fig: rank} and Table~\ref{apx: rank} in Appendix~\ref{sec: parameter_size}, respectively. In Figure~\ref{fig: rank}, after low-rank decomposition ($\text{rank} \neq 0$), the performance in AUC generally improves, and the effects of different ranks are inconsistent across different datasets. In Table~\ref{apx: rank}, the parameter size of the generator increases slightly with the rank, indicating that by adjusting different ranks, an effective balance between the performance and resource consumption of Cuff-KT can be achieved.

\section{Conclusion}
\label{sec:conclusion}

Our paper aims to tackle the RLPA task in KT by proposing a controllable, tuning-free, fast, and flexible method called Cuff-KT to improve adaptability of KT models in real-world scenarios. We decompose the RLPA task to be solved into two sub-issues: intra-learner shift and inter-learner shift, and design a controller to select learners with parameter update values as well as a generator capable of generate personalized parameters based on the current stage or group, thereby achieving adaptive generalization. In instance validations across multiple KT models, Cuff-KT exhibits superior performance in adapting to rapidly changing distributions, avoiding the overfitting and high time overhead challenges inherent in fine-tuning based methods.

\begin{acks}
This research was partially supported by grants from the "Pioneer" and "Leading Goose" R\&D Program of Zhejiang under Grant No. 2025C02022, and the National Natural Science Foundation of China (No.62037001, No.62307032).
\end{acks}

\bibliographystyle{ACM-Reference-Format}
\bibliography{kdd2025}
\appendix

\section{Datasets}
\label{sec: datasets}

The descriptions of the five benchmark datasets used in our experiments are as follows.
\begin{itemize}[leftmargin=*]
    \item \textbf{assist15}\footnote{\url{https://sites.google.com/site/assistmentsdata/datasets/2015-assistments-skill-builder-data}}: The assist15 dataset, is collected from the ASSISTments platform in the year of 2015. It includes a total of 708,631 interactions involving 100 distinct concepts from 19,917 learners.
    
    \item \textbf{assist17}\footnote{\url{https://sites.google.com/view/assistmentsdatamining/dataset?authuser=0}}: The assist17 dataset originates from the 2017 data mining competition and is sourced from the same platform as assist15. It features 942,816 interactions involving 102 concepts from 1,709 learners.
    
    \item \textbf{comp}\footnote{\url{https://github.com/wahr0411/PTADisc}}: The comp dataset, is part of the PTADisc, which encompasses a wide range of courses from the PTA platform. PTADisc includes data from 74 courses, involving 1,530,100 learners and featuring 4,504 concepts, 225,615 questions, as well as an extensive log of over 680 million learner responses. The comp dataset is specifically selected for KT task in Computational Thinking course.
    
    \item \textbf{xes3g5m}\footnote{\url{https://github.com/ai4ed/XES3G5M}}: The xes3g5m dataset incorporates rich auxiliary information such as tree-structured concept relationships, question types, textual contents, and learner response timestamps and includes 7,652 questions and 865 concepts, with a total of 5,549,635 interactions from 18,066 learners. 
    
    \item \textbf{dbe-kt22}\footnote{\url{https://dataverse.ada.edu.au/dataset.xhtml?persistentId=doi:10.26193/6DZWOH}}: Collected from an introduction to relational databases course at the Australian National University, the dbe-kt22 dataset includes data from undergraduate and graduate learners across multiple disciplines between 2018 and 2021. The dataset features detailed meta-data for questions and learning concepts, relationships among concepts and questions, expert-verified question difficulty levels, and various indicators of learner uncertainty.
\end{itemize}

Following the data preprocessing method outlined in~\cite{lee2022contrastive}, we exclude learners with fewer than five interactions and all interactions involving nameless concepts. \textbf{Since a question may involve multiple concepts, we convert the unique combinations of concepts within a single question into a new concept.} Table~\ref{tab:statistics} provides a statistical overview of the processed datasets. It's noted that the large datasets (comp and xes3g5m) are randomly sampled 5000 learners with random seed 12405 due to the heavy training time.

\begin{table}[t]
\centering
\caption{Statistics of five datasets.}
\vspace{-0.4cm}
\label{tab:statistics}
\resizebox{0.5\textwidth}{!}{
\begin{tabular}{@{}ccccc@{}}
\toprule[1.5pt]
\textbf{Datasets}  & \textbf{\#learners} & \textbf{\#questions} & \textbf{\#concepts} & \textbf{\#interactions} \\ 
\midrule
assist15  & 17,115     & 100         & 100        & 676,288        \\
assist17  & 1,708     & 3,162         & 411        & 934,638        \\
comp      & 5,000      & 7,460       & 445        & 668,927        \\
xes3g5m   & 5,000      & 7,242       & 1,221      & 1,771,657      \\
dbe-kt22   & 1,186      & 212       & 127      & 306,904      \\
\bottomrule[1.5pt]
\end{tabular}
}
\end{table}

\section{Baselines}
\label{sec: baselines}
We select five representative KT models with different structures for optimization.

\begin{itemize}[leftmargin=*]
    \item{\textbf{\textcolor[rgb]{1,0,0}{DKT}}: \textcolor[rgb]{1,0,0}{DKT} is a seminal model that leverages a Recurrent Neural Network, specifically utilizing a single layer \textbf{\textcolor[rgb]{1,0,0}{LSTM}}, to directly model learners' learning processes and predict their performance.}
    \item{\textbf{\textcolor[rgb]{1,0.8,0}{DKVMN}}: \textcolor[rgb]{1,0.8,0}{DKVMN} incorporates a Memory-Augmented Neural Network (\textbf{\textcolor[rgb]{1,0.8,0}{MANN}}), utilizing a static key matrix to capture the interconnections among latent concepts and a dynamic value matrix to track and update a student's knowledge mastery in real-time.}
    
    \item{\textbf{\textcolor{blue}{AT-DKT}}: \textcolor{blue}{AT-DKT}, based \textbf{\textcolor{blue}{LSTM and attention}} module, augments the original deep knowledge tracing model by embedding two auxiliary learning tasks: one for predicting concepts and another for assessing individualized prior knowledge. This integration aims to sharpen the model’s predictive accuracy and deepen its understanding of learner performance.}
    \item{\textbf{\textcolor[rgb]{0.0,0.9,0.5}{stableKT}}: \textcolor[rgb]{0.0,0.9,0.5}{stableKT} excels in length generalization, maintaining high prediction performance even with long sequences of student interactions. It uses a multi-head aggregation module combining dot-product and hyperbolic \textbf{\textcolor[rgb]{0.0,0.9,0.5}{attention}} to capture complex relationships between questions and concepts.}
    \item{\textbf{\textcolor[rgb]{0.54,0.17,0.89}{DIMKT}}: \textcolor[rgb]{0.54,0.17,0.89}{DIMKT} uses a \textbf{\textcolor[rgb]{0.54,0.17,0.89}{customized adaptive sequential neural network}} to enhance the assessment of learners' knowledge states by explicitly incorporating the difficulty level of questions and establishes the relationship between learners' knowledge states and difficulty level during the practice process.}
\end{itemize}

The following are the introduction of the three fine-tuning-based methods compared in our experiments.
\begin{itemize}[leftmargin=*]
    \item{\textbf{Full Fine-tuning (FFT)}: FFT involves training all parameters of a model completely. It usually has the highest potential for performance, but it consumes the most resources, takes the longest time to train, and is prone to overfitting when the corpus is not large enough.}
    \item{\textbf{Adapter-based tuning (Adapter)}: Adapter inserts downstream task parameters, known as adapters, into each Transformer block of the pre-trained model. Each adapter consists of two layers of MLP and an activation function, responsible for reducing and increasing the dimensionality of the Transformer's representations. During fine-tuning, the main model parameters are frozen, and only the task-specific parameters are trained. Since the backbone models might not include a Transformer, in our experiments, it is replaced by linear layers.}
    \item{\textbf{Bias-term Fine-tuning (BitFit)}: BitFit is a sparse fine-tuning method that efficiently tunes only the parameters with bias, while all other parameters are fixed. This method tends to be effective on small to medium datasets and can even compete with other sparse fine-tuning methods on large datasets.}
\end{itemize}

\begin{figure*}[t]
    \centering
    \includegraphics[width=0.9\linewidth]{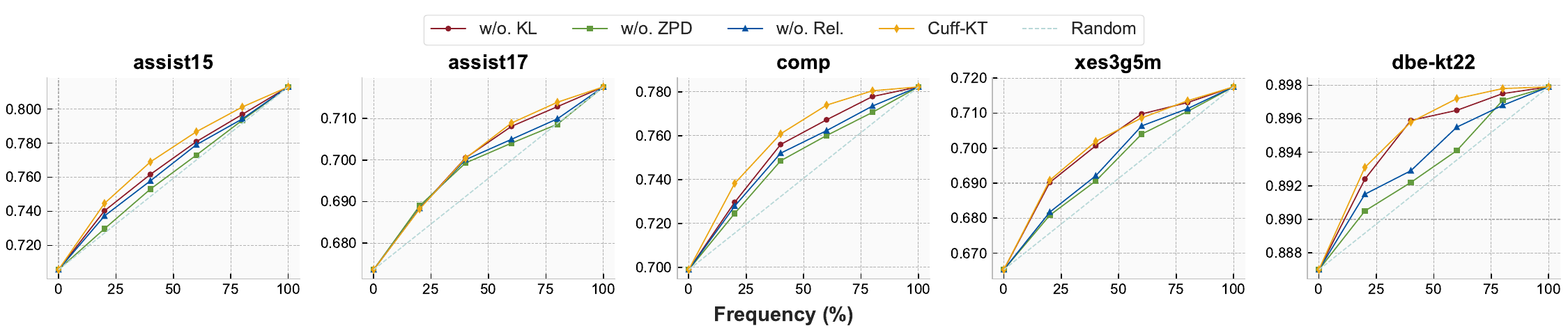}
    \vspace{-0.4cm}
    \caption{Ablation study of the controller in Cuff-KT at different frequencies.}
    \label{apx: control_dkt}
    \vspace{-0.4cm}
\end{figure*}

\begin{figure*}[t]
    \centering
    \includegraphics[width=0.9\linewidth]{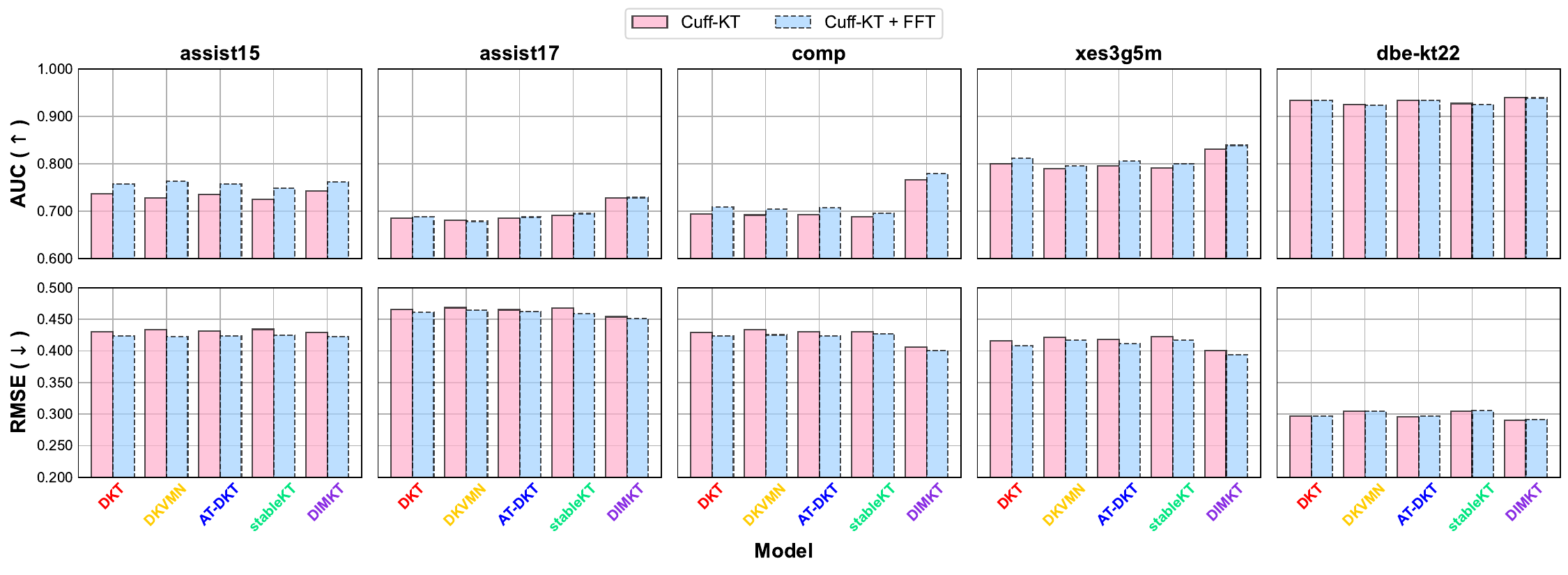}
    \vspace{-0.55cm}
    \caption{The performance of Cuff-KT + FFT under inter-learner shift.}
    \label{apx: cuff+}
    \vspace{-0.3cm}
\end{figure*}

\section{Anomaly Detection Algorithms}
\label{sec: algorithms}

The following four unsupervised anomaly detection algorithms serve as baselines for the controller in Cuff-KT.

\begin{itemize}[leftmargin=*]
    \item{\textbf{LOF}: LOF quantifies the local outlier degree of samples by calculating a score. This score reflects the ratio of the average density of the local neighborhood around a sample point to the density at the location of that sample point. A ratio significantly greater than $1$ indicates that the density at the sample point's location is much lower than the average density of its surrounding neighborhood, suggesting that the point is more likely to be a local outlier.}
    \item{\textbf{PCA}: After performing eigenvalue decomposition, the eigenvectors obtained from PCA reflect different directions of variance change in network traffic data, while eigenvalues represent the magnitude of variance in the corresponding directions. Thus, the eigenvector associated with the largest eigenvalue represents the direction of maximum variance in network traffic data, while the eigenvector associated with the smallest eigenvalue represents the direction of minimum variance. If an individual network connection sample exhibits characteristics inconsistent with the overall network traffic sample, such as deviating significantly from other normal connection samples in certain directions, it may indicate that this connection sample is an outlier.}
    \item{\textbf{IForest}: IForest employs an innovative anomaly isolation method to identify anomalous samples by constructing a binary tree structure (called an Isolation Tree or iTree). Unlike traditional methods, IForest does not build a model of normal samples, but instead directly isolates anomalies. In this process, anomalous samples tend to be isolated more quickly and thus are positioned closer to the root node in the tree, while normal samples are isolated deeper in the tree. By constructing multiple iTrees (typically T trees), the average path length from anomalies to the root node is significantly shorter than that of normal points, and this characteristic is used for anomaly detection. This approach excels in handling large-scale datasets and high-dimensional data, with the advantages of linear time complexity and low memory requirements.}
    \item{\textbf{ECOD}: ECOD is a novel unsupervised anomaly detection algorithm. Its core idea stems from the definition of outliers—typically rare events occurring in the tails of a distribution. The algorithm cleverly uses empirical cumulative distribution functions (ECDF) to estimate the joint cumulative distribution function of the data, thereby calculating the probability of outliers. The uniqueness of ECOD lies in its avoidance of the slow convergence problem of joint ECDF in high-dimensional data. The algorithm calculates the univariate ECDF for each dimension separately, then estimates the degree of anomaly for multidimensional data points through an independence assumption. This is done by multiplying the estimated tail probabilities of all dimensions.}
\end{itemize}

\section{Ablation Study of the Controller in Cuff-KT}
\label{sec: ablation_controller}

\begin{table}
\centering
\caption{The parameter size (k) of the generator with different ranks in Cuff-KT under intra-learner shift.}
\vspace{-0.4cm}
\label{apx: rank}
\resizebox{0.37\textwidth}{!}{
\begin{tabular}{@{}c|c|ccccc@{}}
\toprule[1.5pt]
\multirow{2}{*}{\textbf{Dataset}} & \multirow{2}{*}{\textbf{Backbone}} & \multicolumn{5}{c}{\textbf{Rank}}                     \\ \cmidrule(l){3-7} 
                         &                           & \textbf{0}        & \textbf{1}      & \textbf{2}      & \textbf{3}      & \textbf{4}      \\ \midrule 
                         & DKT                       & 130.12   & 29.96  & 33.22  & 36.49  & 39.75  \\
                         & DKVMN                     & 343.42   & 85.44  & 89.60  & 93.76  & 97.92  \\
assist15                 & AT-DKT                    & 130.12   & 29.96  & 33.22  & 36.49  & 39.75  \\
                         & stableKT                  & 54.98    & 23.26  & 24.32  & 25.38  & 26.43  \\
                         & DIMKT                     & 54.98    & 23.26  & 24.32  & 25.38  & 26.43  \\ \midrule \midrule
                         & DKT                       & 478.75   & 70.08  & 83.29  & 96.51  & 109.72  \\
                         & DKVMN                     & 363.33   & 105.34 & 109.50 & 113.66 & 117.82  \\
assist17                 & AT-DKT                    & 478.75   & 70.08  & 83.29  & 96.51  & 109.72  \\
                         & stableKT                  & 64.93    & 33.22  & 34.27  & 35.33  & 36.38  \\
                         & DIMKT                     & 64.92    & 33.22  & 34.27  & 35.33  & 36.38  \\ \midrule \midrule
                         & DKT                       & 516.86   & 74.46  & 88.77  & 103.07 & 117.37 \\
                         & DKVMN                     & 365.50   & 107.52 & 111.68 & 115.84 & 120.00  \\
comp                     & AT-DKT                    & 516.86   & 74.46  & 88.77  & 103.07 & 117.37 \\
                         & stableKT                  & 66.02    & 34.30  & 35.36  & 36.42  & 37.47  \\
                         & DIMKT                     & 66.02    & 34.30  & 35.36  & 36.42  & 37.47  \\ \midrule \midrule
                         & DKT                       & 1,386.76 & 174.57 & 213.70 & 252.84 & 291.97 \\
                         & DKVMN                     & 415.16   & 157.18 & 161.34 & 165.50 & 169.66  \\
xes3g5m                  & AT-DKT                    & 1,386.76 & 174.57 & 213.70 & 252.84 & 291.97 \\
                         & stableKT                  & 90.85    & 59.14  & 60.19  & 61.25  & 62.30  \\
                         & DIMKT                     & 90.85    & 59.14  & 60.19  & 61.25  & 62.30  \\ \midrule \midrule
                         & DKT                       & 160.38   & 33.44  & 37.57  & 41.70  & 45.82 \\
                         & DKVMN                     & 345.15   & 87.17  & 91.33  & 95.49  & 99.65  \\
dbe-kt22                 & AT-DKT                    & 160.38   & 33.44  & 37.57  & 41.70  & 45.82 \\
                         & stableKT                  & 55.84    & 24.13  & 25.18  & 26.24  & 27.30  \\
                         & DIMKT                     & 55.84    & 24.13  & 25.18  & 26.24  & 27.30  \\ 
                         
                         \bottomrule[1.5pt]
\end{tabular}
}
\vspace{-0.4cm}
\end{table}

We further analyze the influence of different components of the controller in Cuff-KT under intra-learner shift. We instantiate DKT on all datasets. The AUC performance results are shown in Figure~\ref{apx: control_dkt}. We observe that the performance consistently drops the most when the controller removes ZPD (``w/o. ZPD", \textit{i.e.}, without considering coarse-grained changes in knowledge states). This indicates that considering coarse-grained knowledge state changes is crucial, which aligns with reality, as in practical scenarios, a learner's progress or regression is often judged by an overall score. Additionally, when ZPD does not take into account actual length (``w/o. Rel.", \textit{i.e.},  without considering the reliability of ZPD), the performance drops the second most. This is because when a learner has more activity records, their knowledge state is more likely to experience drastic changes, and such learners should receive more attention. On the other hand, when a learner has limited records, the simulated changes in their knowledge state are less reliable and should be given lower weight. When the controller does not consider fine-grained changes in the knowledge state (``w/o. KL"), the performance shows a slight decrease. We attribute this to the fact that when fine-grained knowledge states decline overall, the learner’s knowledge state will experience a major shift. However, such situations are relatively rare in reality.

\section{Performance of Cuff-KT + FFT under Inter-learner Shift}
\label{sec: cuff+_inter}
Figure~\ref{apx: cuff+} shows the performance results of Cuff-KT combined with FFT under inter-learner shift.

\section{Parameter Size of the Generator with Different Ranks in Cuff-KT}
\label{sec: parameter_size}
Table~\ref{apx: rank} shows the parameter sizes of the generator with different ranks in Cuff-KT. It can be observed that low-rank decomposition significantly reduces the parameter size, and increasing the rank only results in a slight increase in the number of parameters.

\end{document}